\pdfoutput=1
\documentclass[10pt,twocolumn,letterpaper]{article}

\usepackage[pagenumbers]{cvpr}

\usepackage{cuted}

\usepackage{currfile}

\usepackage{caption}

\newcommand{\ours}{Agent edit}
\newcommand{\oursF}{Agent formula}

\definecolor{cvprblue}{rgb}{0.21,0.49,0.74}
\usepackage[pagebackref,breaklinks,colorlinks,allcolors=cvprblue]{hyperref}
\usepackage{listings}
\usepackage{pgf}

\title{Gradient-Free Few-Shot Adaptation of Vision Models via Agentic Coding}

\author{Khayrul Islam\\[2pt]
\normalsize Computational Engineering Division, Lawrence Livermore National Laboratory,\\
\normalsize 7000 East Avenue, Livermore, CA 94550, USA}

\usepackage{adjustbox}%
\providecommand{\pmstd}[1]{{\scriptsize$\pm$#1}}

\providecommand{\na}{--}

\newcommand{\gdaFewshotClipEurosatNsixteen}{88.8}%
\newcommand{\opusOvlClipEurosatNsixteen}{77.7}%
\newcommand{\lppFewshotClipPetsNone}{89.8}%
\newcommand{\opusOvlClipPetsNone}{89.8}%
\newcommand{\adamfullAircraftNfiveFoneCellsStd}{2.1}%
\newcommand{\adamfullCubNfiveFoneCellsStd}{4.6}%
\newcommand{\opusOvlCubNfive}{34.6}%
\newcommand{\adamfullPetsNfiveFoneCellsStd}{8.9}%
\newcommand{\opusOvlPetsNfive}{31.5}%
\newcommand{\adamfullCarsNfiveFoneCellsStd}{16.9}%
\newcommand{\opusSamBbbcNtwentyRandom}{0.0}%
\newcommand{\adamSamLivecellNtwenty}{0.0}%
\newcommand{\nEvalHard}{120}%
\newcommand{\nEvalIsaid}{119}%
\newcommand{\nEvalPerClass}{12}%
\newcommand{\budgetPasses}{25}%
\newcommand{\budgetExec}{300}%
\newcommand{\retFtLo}{47.3}%
\newcommand{\retFtHi}{51.4}%
\newcommand{\retTfaLo}{7.6}%
\newcommand{\retTfaHi}{18.8}%
\newcommand{\tfaNovelLo}{0.0}%
\newcommand{\tfaNovelHi}{23.3}%
\newcommand{\oursNovelLo}{22.3}%
\newcommand{\oursNovelHi}{65.0}%
\newcommand{\nWinAgent}{2}%
\newcommand{\vlmShipsAcc}{40.8\%}%
\newcommand{\nEvalFine}{120}%
\newcommand{\agentMinLo}{8.9}%
\newcommand{\agentMinHi}{28.4}%
\newcommand{\agentUsdLo}{1.59}%
\newcommand{\agentUsdHi}{4.57}%
\newcommand{\auditRunsHand}{5}%
\newcommand{\auditRunsTotal}{54}%
\newcommand{\auditRunsFit}{54}%
\newcommand{\cfGpuLo}{75.8}%
\newcommand{\cfGpuHi}{295.2}%
\newcommand{\formCpuLo}{173}%
\newcommand{\formCpuHi}{581}%
\newcommand{\bootTwoLevelResamples}{10000}%
\newcommand{\ciWidthRatio}{2.0}%
\newcommand{\statAgentMeanCub}{11.2}%
\newcommand{\statAgentMeanCubLo}{2.5}%
\newcommand{\statAgentMeanCubHi}{19.5}%
\newcommand{\statAgentMeanPets}{12.3}%
\newcommand{\statAgentMeanPetsLo}{1.0}%
\newcommand{\statAgentMeanPetsHi}{23.3}%
\newcommand{\statFormBestCub}{11.2}%
\newcommand{\statFormBestCubLo}{1.6}%
\newcommand{\statFormBestCubHi}{21.0}%
\newcommand{\statFormBestPets}{11.7}%
\newcommand{\statFormBestPetsLo}{2.0}%
\newcommand{\statFormBestPetsHi}{22.8}%
\newcommand{\statShipsNamed}{0.0}%
\newcommand{\seedWinsAgentCub}{4}%
\newcommand{\seedWinsAgentPets}{4}%
\newcommand{\seedWinsFormBestCub}{5}%
\newcommand{\seedWinsFormBestPets}{5}%
\newcommand{\retAgentAllShots}{0.8}%
\newcommand{\nAgentRunsAll}{88}%
\newcommand{\retFormAllShots}{1.1}%
\newcommand{\nFormRunsAll}{88}%
\newcommand{\retAgentMean}{1.0}%
\newcommand{\retWorst}{5.5}%
\newcommand{\guardMax}{0.56}%
\newcommand{\nAgentRunsGuard}{336}%
\newcommand{\retLwfLo}{40.4}%
\newcommand{\retLwfHi}{48.8}%
\newcommand{\retClassicLo}{0.02}%
\newcommand{\retClassicHi}{1.3}%
\newcommand{\convAll}{166}%
\newcommand{\convTotal}{243}%
\newcommand{\convAfter}{124}%
\newcommand{\convAfterTotal}{195}%
\newcommand{\convExact}{45}%
\newcommand{\ldaCount}{25}%
\newcommand{\nRecipesAudited}{276}%
\newcommand{\nRecipesFlagged}{28}%
\newcommand{\headMsPerImage}{6.4}%
\newcommand{\gateMacroLo}{1.4}%
\newcommand{\gateMacroHi}{4.0}%
\newcommand{\nClassicLose}{11}%
\newcommand{\priorStripMax}{3.7}%
\newcommand{\priorPermLo}{+3.4}%
\newcommand{\priorPermHi}{+4.7}%
\newcommand{\priorHidShufTotal}{8}%
\newcommand{\priorHidShufExcl}{0}%
\newcommand{\priorMixLo}{+0.4}%
\newcommand{\priorMixHi}{+7.5}%
\newcommand{\priorMixExcl}{0}%
\newcommand{\shipsHidProbe}{7.5}%
\newcommand{\shipsHidProbeLo}{4.0}%
\newcommand{\shipsHidProbeHi}{10.0}%
\newcommand{\shipsHidMean}{4.8}%
\newcommand{\shipsHidMeanLo}{1.6}%
\newcommand{\shipsHidMeanHi}{7.6}%
\newcommand{\shipsHidNamed}{4.8}%
\newcommand{\shipsHidNamedLo}{0.3}%
\newcommand{\shipsHidNamedHi}{8.9}%
\newcommand{\shipsHidNamedWon}{3}%
\newcommand{\shipsSeeds}{3}%
\newcommand{\jointAgentAlone}{41.6}%
\newcommand{\jointAgentJoint}{30.5}%
\newcommand{\jointFormAlone}{47.2}%
\newcommand{\jointFormJoint}{42.0}%
\newcommand{\jointCfBestAlone}{43.0}%
\newcommand{\jointCfBestJoint}{34.3}%
\newcommand{\jointFormCars}{65.0}%
\newcommand{\jointCfBestCars}{53.4}%
\newcommand{\jointParentCarsStock}{47.2}%
\newcommand{\jointParentCarsAgent}{46.5}%
\newcommand{\jointParentCarsForm}{44.5}%
\newcommand{\jointFormCub}{29.2}%
\newcommand{\jointCfBestCub}{26.7}%
\newcommand{\jointParentCubStock}{35.7}%
\newcommand{\jointParentCubAgent}{22.7}%
\newcommand{\jointParentCubForm}{30.1}%
\newcommand{\jointFormPets}{53.5}%
\newcommand{\jointCfBestPets}{36.6}%
\newcommand{\jointParentPetsStock}{79.6}%
\newcommand{\jointParentPetsAgent}{46.3}%
\newcommand{\jointParentPetsForm}{53.3}%
\newcommand{\jointFormAircraft}{20.1}%
\newcommand{\jointCfBestAircraft}{20.7}%
\newcommand{\jointParentAircraftStock}{61.4}%
\newcommand{\jointParentAircraftAgent}{42.5}%
\newcommand{\jointParentAircraftForm}{44.0}%
\newcommand{\jointDropHi}{33.3}%
\newcommand{\jointCfParentStock}{65.2}%
\newcommand{\jointCfParentAfter}{6.8}%
\newcommand{\jointHierWorseForm}{44}%
\newcommand{\jointHierRunsForm}{52}%
\newcommand{\tsScoreLo}{32.6}%
\newcommand{\tsScoreHi}{95.3}%
\newcommand{\tsAbove}{5}%
\newcommand{\tsSets}{5}%
\newcommand{\stockCdfsodClipart}{91.2\%}%
\newcommand{\dirOverCapShare}{33\%}%
\newcommand{\cdvReproAp}{40.72}%
\newcommand{\featSets}{5}%
\newcommand{\featNcmBeat}{0}%
\newcommand{\featNcmLose}{1}%
\newcommand{\featOverRows}{2}%
\newcommand{\featUnderRows}{1}%
\newcommand{\lowOracleTested}{3}%
\newcommand{\lowOracleBelow}{1}%
\newcommand{\cdvSeedsFg}{5}%
\newcommand{\cdvSetsTotal}{5}%
\newcommand{\cdvWinsPub}{2}%
\newcommand{\cdvWinsTuned}{2}%
\newcommand{\cdvFormBeatsPub}{3}%
\newcommand{\cdvLateCells}{8}%
\newcommand{\cdvLateWins}{3}%
\newcommand{\cdfsodPubClipart}{41.1}%
\newcommand{\cdfsodOursDropLo}{0.4}%
\newcommand{\cdfsodOursDropHi}{1.2}%
\newcommand{\cdfsodGradDropLo}{7.6}%
\newcommand{\cdfsodGradDropHi}{38.0}%
\newcommand{\jcalCostLo}{4.0}%
\newcommand{\jcalCostHi}{28.0}%
\newcommand{\jcalDropLo}{0.9}%
\newcommand{\jcalDropHi}{5.0}%
\newcommand{\jcalTunedPassed}{1}%
\newcommand{\jcalTunedRuns}{20}%
\newcommand{\noldaFlagged}{30}%
\newcommand{\noldaConvergeFlagged}{25}%
\newcommand{\noldaHeadlineFive}{0}%
\newcommand{\noldaFiveCells}{4}%
\newcommand{\noldaFiveUnchanged}{4}%
\newcommand{\noldaHeadlineTen}{3}%
\newcommand{\noldaTenMove}{3.3}%
\newcommand{\randomEditMax}{0.6}%
\newcommand{\replayCarsOne}{88.3}%
\newcommand{\liveCarsOne}{93.2}%
\newcommand{\replayCarsTwo}{87.5}%
\newcommand{\liveCarsTwo}{92.4}%
\newcommand{\replayXsetLo}{41.3}%
\newcommand{\replayXsetHi}{49.2}%
\newcommand{\liveXsetLo}{54.2}%
\newcommand{\liveXsetHi}{74.8}%
\newcommand{\repeatSdCars}{7.3}%
\newcommand{\repeatSdPets}{6.3}%
\newcommand{\vlmCarsAcc}{99.7\%}%
\newcommand{\vlmCropLo}{80}%
\newcommand{\vlmCropHi}{100}%
\newcommand{\vlmCostLo}{0.0055}%
\newcommand{\vlmCostHi}{0.0267}%
\newcommand{\vlmSecLo}{1.0}%
\newcommand{\vlmSecHi}{3.3}%
\newcommand{\headNameLo}{24}%
\newcommand{\headNameHi}{73}%
\newcommand{\cdvCocoFt}{16.8}%
\newcommand{\cdvCocoStock}{31.7}%
\newcommand{\dirCeiling}{53\%}%
\newcommand{\nSonnetCells}{15}%
\newcommand{\sonnetLo}{1.4}%
\newcommand{\sonnetHi}{10.3}%
\newcommand{\sonnetTrainLo}{2}%
\newcommand{\sonnetTrainHi}{20}%
\newcommand{\opusCellsLo}{18.7}%
\newcommand{\opusCellsHi}{93.2}%
\newcommand{\samNucleiLo}{75.1}%
\newcommand{\samNucleiHi}{79.0}%
\newcommand{\samNucleiAdam}{71.8}%
\newcommand{\samNucleiTrained}{84.9}%
\newcommand{\faVariants}{10}%
\newcommand{\faDropBest}{0.10}%
\newcommand{\faDropAgent}{0.81}%
\newcommand{\faDropForm}{0.90}%
\newcommand{\faParentStock}{65.2}%
\newcommand{\faParentFa}{63.5}%
\newcommand{\faParentAgent}{43.8}%
\newcommand{\faParentForm}{52.2}%
\newcommand{\faParentPetsStock}{92.1}%
\newcommand{\faParentPetsFa}{89.4}%
\newcommand{\faParentPetsAgent}{30.0}%
\newcommand{\faParentPetsForm}{54.6}%
\newcommand{\faFormWinsK}{3}%
\newcommand{\faFormGainKLo}{17.6}%
\newcommand{\faFormGainKHi}{34.5}%
\newcommand{\faAgentWinsK}{3}%
\newcommand{\faAgentGainKLo}{11.3}%
\newcommand{\faAgentGainKHi}{26.6}%
\newcommand{\faFormWinsJ}{3}%
\newcommand{\faAgentWinsJ}{2}%
\newcommand{\faTfaScore}{0.1}%
\newcommand{\faPassedCtwoCut}{0}%
\newcommand{\faPassedCfiveCut}{6}%
\newcommand{\faPassedFa}{20}%
\newcommand{\faPassedFaNeg}{20}%
\newcommand{\faPassedTfaStock}{20}%
\newcommand{\faPassedTotal}{20}%
\newcommand{\wiseScoreLo}{0.0}%
\newcommand{\wiseScoreHi}{1.5}%
\newcommand{\lspDropLo}{26.2}%
\newcommand{\lspDropHi}{43.9}%
\newcommand{\mpFormDiffCars}{+40.1}%
\newcommand{\mpFormLoCars}{+25.6}%
\newcommand{\mpFormHiCars}{+51.6}%
\newcommand{\mpFormSame}{3}%
\newcommand{\apArms}{18}%
\newcommand{\apSpearmanAp}{0.95}%
\newcommand{\solVsOpusLevel}{3}%
\newcommand{\solCocoSolMean}{0.43}%
\newcommand{\solRecipesTotal}{18}%
\newcommand{\freshBestDiffCub}{\ensuremath{-}12.6}%
\newcommand{\freshBestLoCub}{\ensuremath{-}26.0}%
\newcommand{\freshBestHiCub}{\ensuremath{-}0.1}%
\newcommand{\freshTieOrBeat}{3}%
\newcommand{\freshBeatCtwo}{4}%
\newcommand{\freshGuardFailed}{4}%
\newcommand{\freshGuardCells}{20}%
\newcommand{\lowAgentMeanWinsOne}{3}%
\newcommand{\lowFormMeanWinsOne}{2}%
\newcommand{\lowAgentMeanWinsTwo}{3}%
\newcommand{\lowFormBestWinsTwo}{2}%
\newcommand{\lowFormMeanWinsTwo}{3}%
\newcommand{\lowAgentBestWinsThree}{2}%
\newcommand{\lowAgentMeanWinsThree}{3}%
\newcommand{\lowFormMeanWinsThree}{2}%
\newcommand{\Msec}[1]{Sec.~\ref{#1} of the main paper}
\newcommand{\Mtab}[1]{Tab.~\ref{#1} of the main paper}

\newcommand{\mval}[1]{\ifcsname #1\endcsname\csname #1\endcsname\else\na\fi}

\newcommand{\meanthree}[1]{%
  \ifcsname #1Szero\endcsname\ifcsname #1Sone\endcsname\ifcsname #1Stwo\endcsname
    \pgfmathparse{(\csname #1Szero\endcsname+\csname #1Sone\endcsname+\csname #1Stwo\endcsname)/3}%
    \pgfmathprintnumber[fixed,precision=1,zerofill]{\pgfmathresult}%
  \else\na\fi\else\na\fi\else\na\fi}
\begin{document}
\maketitle

\begin{abstract}
Adapting a vision model to new classes from a few labeled images usually relies on gradient fine-tuning, which is unstable with so few images.
We ask whether a coding agent can replace gradient descent for this task.
We propose the \ours{}, a method in which a coding agent writes a program that sets the new class weights in the class head of a frozen COCO-trained detector, with no gradient step.
The agent runs the detector on the labeled images a limited number of times and sees how much its edit lowers COCO mAP; the program rebuilds the weights without the agent.
In the same head and with the same labels, the \ours{} outperforms full-decoder fine-tuning at five shots on \nWinAgent{} of four fine-grained sets, by \statAgentMeanCub{} and \statAgentMeanPets{} F1 points, and ties on the rest.
At ten shots, fine-tuning ties with or outperforms it.
Moreover, when the new classes are scored jointly with the COCO classes, the AP of the COCO class to which they belong falls by up to \jointDropHi{} points, and limiting that loss lowers F1 on the new classes.
Most of the agent's programs share three parts, which we write as a formula that needs no agent.
A classifier on frozen backbone features reaches a higher F1 than every method inside the head on every set.
Hence a coding agent can replace gradient descent only in part: inside the head and at five shots.
Code is available at \url{https://github.com/Khayrulbuet13/agent-edit}.
\end{abstract}
\section{Introduction}
\label{sec:intro}

\begin{figure*}[t]
  \centering
  \includegraphics[width=0.50\textwidth]{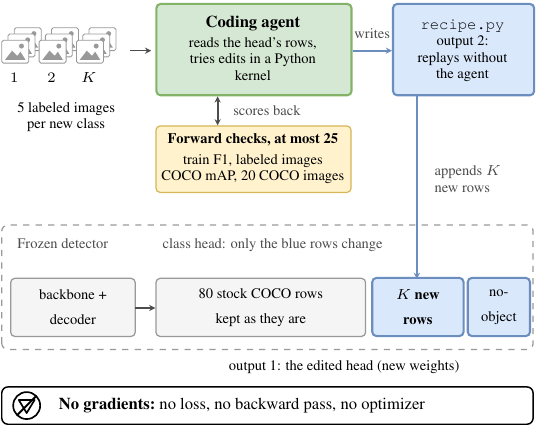}
  \hfill
  \includegraphics[width=0.42\textwidth]{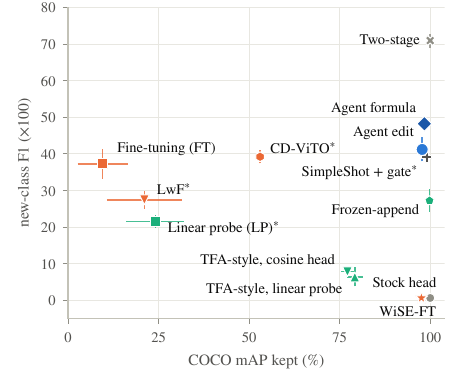}
  \caption{\textbf{A coding agent writes the new class rows of a detector's head.}
  \emph{Left:} the agent runs the model on five labeled images per class within a request budget, checks its edit on COCO, and writes it as a program. No gradient is computed.
  \emph{Right:} new-class F1 against the share of COCO mask mAP kept, at five shots, mean over four fine-grained sets (baselines: Sec.~\ref{sec:setup}).
  Two-stage is the outside-the-head reference, a classifier on frozen features; CD-ViTO is scored on boxes.
  Mean and std over seeds 0--2. $^\ast$ 200-image COCO protocol.}
  \label{fig:teaser}
\end{figure*}

Object detectors are trained on a fixed set of classes, yet users often need classes outside that set and can label only a few images of each.
The standard approach is fine-tuning, which trains the detector, or only its last layers, by gradient descent on the new examples~\cite{wang2020tfa,sun2021fsce,qiao2021defrcn}.
With so few examples, however, fine-tuning is unstable and causes forgetting, a loss of accuracy on the original classes: in our experiments, fine-tuning the decoder lowers mask mAP on the 80 COCO~\cite{coco} classes by \retFtLo--\retFtHi{} points (Sec.~\ref{sec:res-forget}).

Continual-learning methods limit forgetting through distillation, a weight penalty or stored exemplars~\cite{li2016lwf,kirkpatrick2017ewc,li2018l2sp,rebuffi2017icarl}, but they still take gradient steps.
Closed-form methods take no gradient step: they write the weights of a new class directly, e.g., as the mean embedding of its labeled examples~\cite{snell2017prototypical,qi2018imprinted,ganea2021imtfa}.
On strong frozen embeddings, such heads are hard to beat~\cite{wang2019simpleshot}.
However, a closed-form rule is fixed in advance and does not adapt to the model's errors.

Coding agents offer a third possibility: an agent can inspect the weights of a network, run it on labeled images, and express a weight edit as a program.
So far, agents have been evaluated on writing training code~\cite{jiang2025aide,chan2024mlebench,huang2023mlagentbench}, where the weights still come from gradient descent.
Concurrent work~\cite{sparks2026} lets a language model edit the weights of small networks without gradients, but it edits no vision model and has no gradient baseline on the same labels or measurement of forgetting.
It therefore remains open whether an agent-written edit can replace gradient descent when the class head of a detector must learn new classes from a few images, and how much forgetting such an edit causes.

To this end, we propose a harness, software in which a coding agent writes the class rows of the new classes as a program, with no gradient step (Fig.~\ref{fig:teaser}, left).
A class row is the weight vector with which the head scores one class.
The backbone and mask decoder stay frozen.
The agent may run the model on the labeled images a limited number of times and sees how much each edit lowers COCO mAP, so that it can limit forgetting.
The agent returns the edited weights and the program that rebuilds them without the agent.
We call this method the \ours{}.
For a fair comparison, every baseline receives the same labels.
Most of the agent's programs contain a class-mean rule, a term that separates objects from background, and a step that removes shared directions.
We write one closed form of them as the \oursF{} (Eq.~\ref{eq:formula}), which needs no agent.

We evaluate on PMT~\cite{pmt}, a mask transformer on a frozen DINOv3~\cite{dinov3}, with Claude Opus~5.5 as the agent, on four fine-grained detection sets and one satellite set of ships.
With five labeled images per class (five shots), the agent replaces gradient descent only in part.
The \ours{} outperforms full-decoder fine-tuning in F1 on \nWinAgent{} of the four fine-grained sets and ties on the rest, and it lowers COCO mAP by \retAgentMean{} points on average (\retWorst{} in the worst run); the gain disappears by ten shots (Secs.~\ref{sec:res-main} and~\ref{sec:res-shots}).
However, a mean over classes hides the loss of one class: when the new rows and the COCO rows share one head, the AP of the COCO class to which the new classes belong (e.g., bird for gulls and terns) falls by up to \jointDropHi{} points, and gradient descent on the new rows alone preserves it better (Sec.~\ref{sec:res-forget}).
Also, a classifier outside the head, on frozen backbone features, reaches a higher F1 than every method inside the head on every set (gray in Fig.~\ref{fig:teaser} and Tab.~\ref{tab:main}), so editing the head is not the best way to add classes.

In summary, we make three contributions:
\begin{enumerate}\itemsep1pt
\item \textbf{\texttt{Harness and Protocol}}: A harness for gradient-free head editing by a coding agent, with gradient baselines on the same labels, a joint head of new and COCO rows, statistics over seeds and test images, and tests of whether the agent's prior knowledge of the classes explains the result (Secs.~\ref{sec:method}, \ref{sec:setup} and~\ref{sec:res-prior}).
\item \textbf{\texttt{Low-Shot Findings}}: Empirical findings on how the \ours{} compares with gradient fine-tuning at low shot, and on what it costs the old classes (Sec.~\ref{sec:results}).
\item \textbf{\texttt{Program Analysis}}: An analysis of the programs that the agent wrote and of the formula they share, with the formula and the harness released as code (Sec.~\ref{sec:res-program}).
\end{enumerate}
\section{Related work}
\label{sec:related}

\noindent\textbf{Few-shot detection and forgetting.}
TFA~\cite{wang2020tfa} fine-tunes the last layers of a detector on the novel classes. FSCE~\cite{sun2021fsce} and DeFRCN~\cite{qiao2021defrcn} do the same. Meta-learned detectors condition on support images~\cite{kang2019fsrw,yan2019metarcnn}. Generalized and incremental few-shot detection score base and novel classes together and target forgetting~\cite{fan2021retentive,guirguis2022cfa,guirguis2023niff,ma2023digeo,perezrua2020once,yin2022sylph,nguyen2022ifsrcnn}. Cross-domain few-shot detection starts from a pretrained detector~\cite{xiong2022cdfsod,lee2022mofsod}. CD-ViTO~\cite{fu2024cdvito} fine-tunes DE-ViT~\cite{zhang2023devit}. Newer methods simulate the target domain~\cite{zhuo2026psp} or tune the decoder~\cite{yu2026ftfsod}. Continual learning limits drift with distillation~\cite{li2016lwf}, a weight penalty~\cite{kirkpatrick2017ewc,li2018l2sp}, or stored exemplars~\cite{rebuffi2017icarl}. LP-FT~\cite{kumar2022lpft} and WiSE-FT~\cite{wortsman2022wiseft} protect pretrained features, and task arithmetic~\cite{ilharco2023taskarithmetic}, model soups~\cite{wortsman2022modelsoups}, and TIES-Merging~\cite{yadav2023ties} merge fine-tuned weights. Our agent differs in taking no gradient step.

\noindent\textbf{Class-mean heads.}
The closest prior work to the \oursF{} is iMTFA~\cite{ganea2021imtfa}, which appends class-mean embeddings of the support objects to a frozen classifier with no training. The \oursF{} also appends class-mean rows but adds a gate for the one query per object, removal of the main within-class directions~\cite{mu2018allbutthetop}, and a selection of its settings under a limit on the COCO drop. Prototypical networks~\cite{snell2017prototypical} classify by distance to class means. Imprinted weights~\cite{qi2018imprinted} and dynamic few-shot learning~\cite{gidaris2018dynamic} write new class rows into a classifier and keep the old ones. ROME~\cite{meng2022rome} and MEMIT~\cite{meng2023memit} write facts into language-model layers in closed form, and Bau et al.~\cite{bau2020rewriting} and Santurkar et al.~\cite{santurkar2021editing} rewrite one rule of a generative model or a classifier. SimpleShot~\cite{wang2019simpleshot} shows that a simple head on a good frozen embedding is hard to beat. Baseline++~\cite{chen2019closerlook} and Tian et al.~\cite{tian2020rethinking} agree, which our outside-the-head classifier confirms. Deep SLDA, FeCAM, and RanPAC bring class statistics to continual learning~\cite{hayes2020slda,goswami2023fecam,mcdonnell2023ranpac}. Tip-Adapter~\cite{zhang2022tipadapter} fits a CLIP head in closed form. GDA~\cite{wang2024gda} does the same; LP++~\cite{huang2024lpp} trains a linear probe.

\noindent\textbf{Coding agents.}
AIDE~\cite{jiang2025aide}, MLE-bench~\cite{chan2024mlebench}, and MLAgentBench~\cite{huang2023mlagentbench} score agents on writing training code. AlphaEvolve~\cite{novikov2025alphaevolve}, FunSearch~\cite{romeraparedes2024funsearch}, and OPRO~\cite{yang2024opro} search over programs or prompts with an evaluator. In contrast, the program our agent writes sets the weights directly. Concurrent work~\cite{sparks2026} lets an LLM read and edit the weights of small networks and of a language model without gradients; it edits no vision model and has no dense prediction, no trained baselines on the same labels, and no forgetting measurement. Neural functionals~\cite{zhou2023pnf,zhou2023nft,zhou2024unf}, hypernetworks~\cite{charakorn2025texttolora} and Weight Oracles~\cite{kabra2026weightoracles} train a model to write or read weights. MLLMs classify from in-context examples without changing any weights~\cite{huang2024pathology}.
\section{Method}
\label{sec:method}

\begin{figure}[tb]
  \centering
  \includegraphics[width=\linewidth]{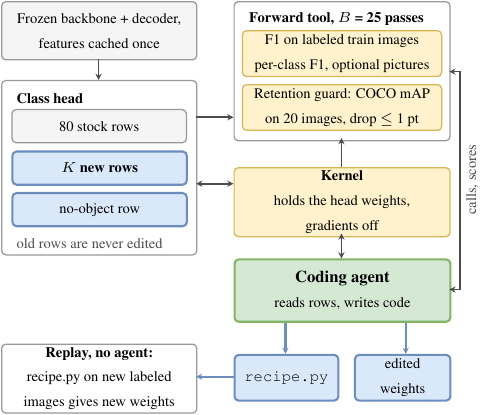}
  \caption{\textbf{The harness.}
  The agent edits the class rows in a Python kernel and reaches the frozen model only through the forward tool.
  The COCO guard checks COCO mAP on 20 images.}
  \label{fig:method}
\end{figure}

\subsection{Problem setting}
\label{sec:method-setup}

PMT~\cite{pmt} is a mask transformer on a frozen DINOv3 ViT-L~\cite{dinov3}.
For each image the frozen decoder returns 200 queries, each with a fixed mask, a fixed mask score and an embedding $q\in\mathbb{R}^{1024}$.
A class head maps $q$ to logits $Wq+b$ over the class rows and one no-object row, the score for background.
A prediction's score is its softmax class probability times the mask score, so an edit can relabel or drop a mask but cannot create one.

A new task has $K$ classes and $N$ labeled training images per class.
The editable set holds the $K$ new rows and the no-object row, and while the agent works the task is scored with these $K{+}1$ rows alone.
For deployment and for the COCO guard, the new rows are appended to the 80 stock rows, the class rows of the pretrained head, and the edited no-object row replaces the stock one; the 80 stock rows stay untouched, and Sec.~\ref{sec:res-forget} also scores the new classes inside that 90-row head.
Every new row starts as a stock row, and one of them holds the task's parent class, the COCO class closest to the task's classes (car, airplane, boat, bird or cat); for birds and cats we copy that row into the first new row, since the first ten stock rows lack it.
SAM~3~\cite{sam3} (Sec.~\ref{sec:res-program}) and CLIP~\cite{clip} (supplement~I) use the same harness.

\subsection{Agent harness}
\label{sec:method-harness}

\noindent\textbf{Tools.}
The agent works in a Python kernel that holds the editable tensors with gradients off (Fig.~\ref{fig:method}), and it may run code at most $C=\budgetExec$ times.
A \emph{generate} request to the forward tool runs the original head, the edited head or both on training images and returns macro-F1 and per-class F1.
A \emph{read} request returns query embeddings and class logits, each marked as matched to a true object or not.
Any request costs one of the $B=\budgetPasses$ passes (the budget).

\noindent\textbf{Rules and label use.}
In the prompt, we ban backward passes, autograd and optimizers, iterative fitting of weights to a loss or to labels, and closed-form regression or least-squares fits of weights to labels (the fitting rule).
We allow statistics that are not fits to labels and edits the agent can explain: shift a bias, remove a direction, set a value from a measured number.
We add a static check that rejects gradient and network imports, and we log every code block and audit the logs for the fitting rule.
Some programs use an LDA-style metric, the inverse of the feature covariance pooled over classes.
It is a closed-form statistic that the fitting rule does not clearly separate from a fit.
Of the \convTotal{} independent programs (from runs that began without another run's program), \ldaCount{} use it and remain in every table (Sec.~\ref{sec:res-program}).
On label use, the agent picks between edits by training F1, and every baseline receives the same labels.

\noindent\textbf{COCO guard.}
The COCO guard scores COCO mask mAP on the first 20 val2017 images for the stock head with the agent's rows appended, and the agent reads it with a retain request, which costs one pass.
We tell the agent to keep the drop within one point, and the harness reports the drop but rejects nothing.

\subsection{From programs to a formula}
\label{sec:method-out}

\noindent\textbf{Weights and program.}
We score the agent's saved tensors once on held-out images.
The agent also writes \texttt{recipe.py}, a function \texttt{edit(train, weights)} that rebuilds the edit from its read outputs and the stock weights; we check that it returns the saved edit on the run's own data.

\noindent\textbf{Agent formula.}
We read the agents' programs for cars, birds and cats, wrote their shared structure as code, and replaced the hand-picked constants with three hyperparameters.
The \emph{primary} query of a labeled object is the matched query the stock head finds most object-like; $\bar q$ is their mean and $\mu_k$ the mean for class $k$, with $\bar\mu$ the mean of the $\mu_k$.
The gate direction $v$ points from the mean of the other queries to $\bar q$, and $\tau$ is the midpoint of the two means' projections on $v$.
The base row $(w_0,b_0)$ is the no-object row for birds and cats; for cars, aircraft and ships, cross-validation chooses between the COCO parent row and the no-object row.
Each new row adds a gate term and a class term to it:
\begin{align}
  w_k &= w_0 + g\,v + s\,d_k, \label{eq:formula}\\
  b_k &= b_0 - g\,\tau - s\big(d_k^{\top}\bar q + \tfrac12\lVert d_k\rVert^2\big) - c, \nonumber
\end{align}
where $d_k=P_m(\mu_k-\bar\mu)$ and $P_m$ removes the top $m$ principal directions of the within-class spread of the primary embeddings~\cite{mu2018allbutthetop}.
The gate term, the part of the edit that decides whether an object is present, raises the one query per object above the no-object row.
The class term is a nearest-mean score~\cite{snell2017prototypical}.
The hyperparameters are the gate gain $g$, the class gain $s$ and $m$, and a bias offset $c$ comes from an F1 sweep on the training images.
We tune them by 5-fold cross-validation F1 on the training images and keep the best setting whose 20-image COCO drop is within one point, the agent's bound.
The \oursF{} needs no agent or GPU; supplement~F gives it as code.
\section{Experiments}
\label{sec:results}

\subsection{Setup}
\label{sec:setup}

\noindent\textbf{Datasets.}
We use four fine-grained detection sets with ten look-alike classes each: sedan models from Stanford Cars~\cite{stanfordcars}, gulls and terns from CUB-200-2011~\cite{cub}, cat breeds from Oxford-IIIT Pet~\cite{pets}, and Boeing~737 and Airbus A319/A320 variants from FGVC-Aircraft~\cite{aircraft}.
The fifth set holds ten vessel types in satellite chips from xView~\cite{lam2018xview}, where a vision-language model classifies only \vlmShipsAcc{} of the vessel crops correctly from the class names alone.
Every set has \nEvalPerClass{} held-out images per class (\nEvalFine{} per set), and training images are drawn from the rest by seed.
We use 1, 2, 3, 5, 10 and 20 shots, except ships (5 shots; its rarest class has five training images).
We run five seeds at 5 and 10 shots and three below 5 shots, at 20 shots and on ships.
Sets with boxes only are scored with box IoU, pets with mask IoU.

\noindent\textbf{Agents.}
Claude Opus~5.5 is the main agent and GPT~6.1 Sol the second (three seeds), with the same prompt; Sonnet~5 is in Sec.~\ref{sec:limits}.
A run has $B=\budgetPasses$ passes and $C=\budgetExec$ code runs.
The main runs are Opus at 5, 10 and 20 shots.

\noindent\textbf{Baselines.}
All baselines use the same frozen features and labeled images as the agent, with a grid selected by 5-fold cross-validation on the training images.
\emph{Gradient, images only:} a linear probe (LP), an imprinted cosine head~\cite{qi2018imprinted}, LwF~\cite{li2016lwf}, and full-decoder fine-tuning (FT)~\cite{kumar2022lpft}, with FT's grid widened until its optimum lies inside and three batch orders per seed (we call the average F1 over the three the mean batch order, and the highest test F1 per seed the best batch order, a gray row).
\emph{Anti-forgetting:} a TFA-style joint head~\cite{wang2020tfa} that also trains on five COCO images per class, and WiSE-FT~\cite{wortsman2022wiseft} and L2-SP~\cite{li2018l2sp} on FT, each tuned under the COCO guard.
\emph{Frozen-append, the matched gradient baseline:} gradient descent in the same head, with the 80 stock rows and the no-object row frozen so only the $K$ new rows train, its settings selected by cross-validation under the 20-image COCO guard.
It has what the agent has, rows that cannot move the old classes and the COCO guard; we report the best of \faVariants{} variants per set, some of which relax the freeze (supplement~B).
\emph{Published:} CD-ViTO~\cite{fu2024cdvito} at its published configuration on \cdvSeedsFg{} seeds, and a tuned variant.
\emph{Closed form:} nearest class mean~\cite{snell2017prototypical}, SimpleShot~\cite{wang2019simpleshot} and shrinkage LDA, given Eq.~\ref{eq:formula}'s gate.
\emph{Outside the head (gray):} a classifier on frozen backbone features that assigns a class to each box of the stock detector, best of seven variants per set, and an agent-written linear head on the same features.
However, four rows are selected with test labels (FT at its best batch order, frozen-append, the outside-the-head classifier, the CD-ViTO variant), which favors the baselines, as do hyperparameters the protocol leaves open.

\noindent\textbf{Metrics.}
We report instance macro-F1 at IoU~0.5, with predictions of score at least 0.5 matched greedily to unused true instances and F1 averaged over classes.
We report AP over IoU 0.5 to 0.95 (box AP on pets for every method), and AP50 and top-1 localization in supplement A.
Forgetting is the drop in COCO mask mAP over the 80 stock classes when the new rows are appended, measured on COCO images no method used for selection (not the 20 guard images) unless marked $^\ast$.
We give the mean over runs and the worst single run.
We score the new classes inside the 90-row head, where the 80 stock rows compete with the new rows, together with the AP of the task's COCO parent class; the new-class F1 in this head is the joint F1.

\noindent\textbf{Statistics.}
We average over seeds.
The seeds share one test set. Thus paired differences, pooled over seeds and images, use a two-level bootstrap: each of \bootTwoLevelResamples{} resamples draws seeds with replacement, then one set of test images used for every drawn seed.
We give the 95\% percentile interval and the number of seeds where the difference is positive, since five seeds allow no exact seed-level test below $p=0.0625$.
A bootstrap over test images alone gives intervals \ciWidthRatio{} times narrower (supplement~A).
Where the two-level interval includes zero, we call the comparison a tie.
All results in this section use PMT on a frozen DINOv3 with Claude Opus~5.5 as the agent, unless stated otherwise, and F1 is instance macro-F1 at IoU~0.5 multiplied by 100.

\subsection{Comparison with gradient descent}
\label{sec:res-main}
\label{sec:res-cdfsod}

\begin{table*}[t]
  \centering
  \small
\begingroup\setlength{\tabcolsep}{3pt}
\begin{adjustbox}{max width=\linewidth}
\begin{tabular}{lccccccccc}
\toprule
 & \multicolumn{5}{c}{F1 $\times$100, five shots} & \multicolumn{2}{c}{Mean over 4 sets} & \multicolumn{2}{c}{COCO points lost} \\
\cmidrule(lr){2-6}\cmidrule(lr){7-8}\cmidrule(lr){9-10}
Method & Cars & CUB & Pets & Aircraft & Ships & F1 & AP & mean & worst run \\
\midrule
Stock head & 0.6 & 0.9 & 0.0 & 1.2 & 1.1 & 0.7 & 3.1 & 0.0 & 0.0 \\
\midrule
\multicolumn{10}{l}{\scriptsize\emph{Gradient, project images only}} \\
Linear probe (LP) & 30.6\pmstd{4.0} & 15.2\pmstd{3.7} & 17.4\pmstd{3.6} & 22.1\pmstd{6.3} & 10.0\pmstd{0.2} & 21.3 & 22.9 & 42.7$^\ast$ & 56.0$^\ast$ \\
Cosine head & 33.0\pmstd{4.5} & 19.2\pmstd{3.8} & 14.8\pmstd{1.7} & 20.2\pmstd{2.9} & 9.5\pmstd{1.0} & 21.8 & 25.8 & \na & \na \\
Fine-tuning (FT), mean of 3 orders & 58.1\pmstd{13.4} & 22.5\pmstd{6.9} & 39.4\pmstd{3.0} & 20.9\pmstd{2.3} & \textbf{12.4}\pmstd{0.9} & 35.2 & 37.3 & 49.1 & 55.8 \\
\textcolor{gray}{FT, best order$^\ddagger$} & \textcolor{gray}{71.2\pmstd{11.5}$^\ddagger$} & \textcolor{gray}{28.4\pmstd{7.4}$^\ddagger$} & \textcolor{gray}{48.6\pmstd{7.4}$^\ddagger$} & \textcolor{gray}{23.9\pmstd{1.7}$^\ddagger$} & \textcolor{gray}{14.8\pmstd{2.2}$^\ddagger$} & \textcolor{gray}{43.0} & \textcolor{gray}{42.4} & \textcolor{gray}{49.1} & \textcolor{gray}{55.8} \\
FT + LwF & 18.3\pmstd{8.5} & 22.9\pmstd{3.4} & 49.9\pmstd{5.1} & 18.7\pmstd{3.4} & \na & 27.5 & 31.5 & 44.4$^\ast$ & 56.0$^\ast$ \\
\midrule
\multicolumn{10}{l}{\scriptsize\emph{Gradient, with anti-forgetting}} \\
TFA-style, linear probe & 20.8\pmstd{3.5} & 0.0\pmstd{0.0} & 4.9\pmstd{6.9} & 0.0\pmstd{0.0} & \na & 6.4 & 26.1 & 11.3 & 20.2 \\
TFA-style, cosine head & 23.3\pmstd{4.1} & 0.5\pmstd{0.7} & 1.0\pmstd{1.3} & 6.6\pmstd{7.3} & \na & 7.9 & 29.1 & 12.4 & 18.6 \\
TFA from stock weights & 0.0\pmstd{0.0} & 0.0\pmstd{0.0} & 0.3\pmstd{0.6} & 0.3\pmstd{0.6} & \na & 0.1 & 19.9 & 1.3 & 2.1 \\
Frozen-append, best per set$^\ddagger$ & 40.0\pmstd{4.8} & 21.2\pmstd{6.0} & 24.7\pmstd{3.6} & 22.7\pmstd{5.5} & \na & 27.1 & 28.7 & \textbf{0.1} & \textbf{0.6} \\
WiSE-FT & 1.2\pmstd{0.2} & 0.0\pmstd{0.0} & 0.0\pmstd{0.0} & 1.5\pmstd{0.2} & \na & 0.7 & 8.6 & 1.3 & 2.8 \\
L2-SP & 30.8\pmstd{1.2} & 10.8\pmstd{5.7} & 11.0\pmstd{9.3} & 19.8\pmstd{3.4} & \na & 18.1 & \na & 37.3 & 52.7 \\
\midrule
\multicolumn{10}{l}{\scriptsize\emph{Published}} \\
CD-ViTO$^\dagger$ & 42.9\pmstd{3.2}$^\dagger$ & 30.1\pmstd{2.5}$^\dagger$ & 67.1\pmstd{3.0}$^\dagger$ & 14.9\pmstd{1.9}$^\dagger$ & 18.0\pmstd{2.3}$^\dagger$ & 38.7 & 32.0 & \na & \na \\
\midrule
\multicolumn{10}{l}{\scriptsize\emph{Closed-form head, gate of Eq.~\ref{eq:formula}}} \\
SimpleShot + gate & 44.0\pmstd{6.4} & 34.6\pmstd{5.0} & 52.2\pmstd{1.9} & \textbf{25.6}\pmstd{2.3} & \na & 39.1 & 40.9 & 0.5$^\ast$ & 2.5$^\ast$ \\
\midrule
\multicolumn{10}{l}{\scriptsize\emph{Outside the head}} \\
\textcolor{gray}{Two-stage (stock boxes, frozen classifier)} & \textcolor{gray}{95.3\pmstd{0.2}} & \textcolor{gray}{66.4\pmstd{4.0}} & \textcolor{gray}{90.6\pmstd{1.3}} & \textcolor{gray}{32.6\pmstd{5.0}} & \textcolor{gray}{30.4\pmstd{2.3}} & \textcolor{gray}{71.2} & \textcolor{gray}{65.7} & \textcolor{gray}{0.0} & \textcolor{gray}{0.0} \\
\textcolor{gray}{\ours{} on features (stock boxes)} & \textcolor{gray}{74.7\pmstd{5.9}} & \textcolor{gray}{35.7\pmstd{4.2}} & \textcolor{gray}{87.2\pmstd{2.3}} & \textcolor{gray}{16.1\pmstd{2.8}} & \textcolor{gray}{27.4\pmstd{2.9}} & \textcolor{gray}{53.4} & \na & \textcolor{gray}{0.0} & \textcolor{gray}{0.0} \\
\midrule
\multicolumn{10}{l}{\scriptsize\emph{Ours}} \\
\ours{} (Opus) & 58.1\pmstd{12.6} & 32.4\pmstd{5.1} & 52.3\pmstd{11.2} & 23.6\pmstd{4.3} & 11.7\pmstd{2.5} & 41.6 & 44.7 & 1.0 & 5.5 \\
\ours{} (Sol) & 41.0\pmstd{4.5} & 34.0\pmstd{4.3} & 52.4\pmstd{3.9} & 22.2\pmstd{2.5} & \na & 37.4 & 40.1 & 0.4 & 2.6 \\
\oursF{} & \textbf{65.0}\pmstd{5.6} & \textbf{39.6}\pmstd{3.7} & \textbf{61.7}\pmstd{4.1} & 22.3\pmstd{2.9} & 11.2\pmstd{2.1} & \textbf{47.2} & \textbf{50.7} & 1.0 & 3.4 \\
\bottomrule
\end{tabular}
\end{adjustbox}
\par\vspace{2pt}{\scriptsize Mean$\pm$std over seeds: five on each fine-grained set, except Cosine head (3), FT + LwF (3), TFA-style, linear probe (3), TFA-style, cosine head (3), WiSE-FT (3), L2-SP (3), SimpleShot + gate (3), \ours{} on features (stock boxes) (3), \ours{} (Sol) (3). Ships: 3 seeds. $^\ddagger$ Best batch order per seed (FT) or best of 10 guarded variants per set (frozen-append), picked on the test images. WiSE-FT and L2-SP: setting picked under the formula's 20-image guard, which L2-SP meets in 0 of 36 cells (smallest-drop fallback); COCO over every (set, seed, batch order) run. Frozen-append and TFA from stock: COCO on val images 201 to 400. AP: IoU 0.5 to 0.95, box AP on pets for every row. \na{}: not run.}
\endgroup
  \caption{\textbf{Five-shot results on PMT.}
  F1 ($\times$100) per set, mean F1 and AP over the fine-grained sets, and COCO mask mAP points lost when the new rows join the 80 stock rows (mean, then worst run).
  Bold: best non-stock in-head mask-scored value per column (lowest for COCO).
  Gray rows are outside this comparison: FT at its best batch order ($^\ddagger$) and the outside-the-head rows, stock boxes labeled by a frozen-backbone classifier or an agent-written linear head.
  $^\dagger$ Box-IoU scorer; $^\ast$ 200-image COCO protocol (first 20 images: the guard's).}
  \label{tab:main}
\end{table*}

In Tab.~\ref{tab:main}, against full-decoder fine-tuning averaged over batch orders (FT), the \ours{} outperforms FT on CUB by \statAgentMeanCub{} F1 points (95\% interval \statAgentMeanCubLo--\statAgentMeanCubHi) and on pets by \statAgentMeanPets{} (\statAgentMeanPetsLo--\statAgentMeanPetsHi), in \seedWinsAgentCub{} and \seedWinsAgentPets{} of 5 seeds, respectively, and ties on cars and aircraft.
For birds and cats the agent starts from the COCO bird or cat row (Sec.~\ref{sec:method-setup}).
The \oursF{} outperforms the best FT batch order, which we pick using the test labels, on CUB by \statFormBestCub{} F1 points (\statFormBestCubLo--\statFormBestCubHi) and on pets by \statFormBestPets{} (\statFormBestPetsLo--\statFormBestPetsHi), in \seedWinsFormBestCub{} and \seedWinsFormBestPets{} of 5 seeds, respectively, and ties on cars and aircraft.
The ranking does not depend on the 0.5 score threshold: mean F1 and mean AP over \apArms{} methods have a Spearman correlation of \apSpearmanAp{}, and no five-shot gain turns into a loss under AP, AP50 or top-1 localization, although the \oursF{}'s gain over FT's best order on pets becomes a tie under AP.
GPT~6.1 Sol writes the same kind of program, and all \solRecipesTotal{} of its programs replay exactly; its \ours{} ties with Opus's on \solVsOpusLevel{} sets, scores lower F1 on cars, and loses \solCocoSolMean{} COCO points on average.

The gray rows are a reference, not part of this comparison: a stock detector whose boxes a frozen DINOv3 classifier labels scores higher F1 than the \ours{} and the \oursF{} on \tsAbove{} of \tsSets{} sets, with no extra backbone pass, and it cannot forget.
Editing the head is therefore not the highest-F1 way to add these classes, and our result holds inside the class head only.
CD-ViTO, a published detector with another backbone, exceeds both in mean F1 on \cdvWinsPub{} of \cdvSetsTotal{} sets (\cdvWinsTuned{} when tuned on test labels), the \oursF{} outperforms it on \cdvFormBeatsPub{} (one-level interval), and at 10 or 20 shots its mean F1 exceeds both in \cdvLateWins{} of \cdvLateCells{} cells.
On ArTaxOr and DIOR~\cite{xiong2022cdfsod} our edits lose \cdfsodOursDropLo--\cdfsodOursDropHi{} COCO points, against \cdfsodGradDropLo--\cdfsodGradDropHi{} for unguarded heads (supplement J).

\subsection{Effect of the number of shots}
\label{sec:res-shots}

\begin{figure*}[t]
  \centering
  \includegraphics[width=0.94\textwidth]{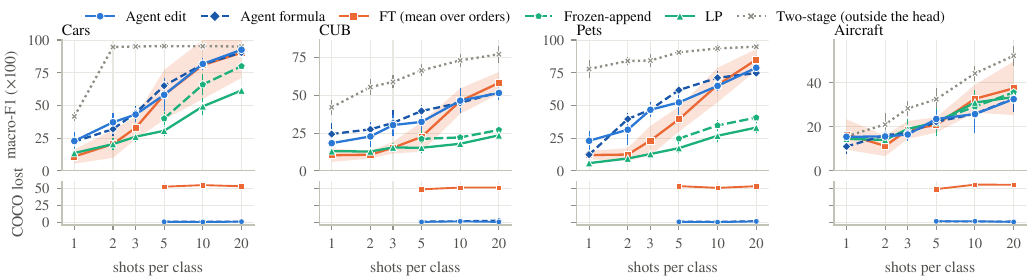}
  \caption{\textbf{F1 against the number of shots.}
  One panel per fine-grained set. Lines: \ours{}, \oursF{}, FT (mean over three batch orders; band: $\pm$1 std over seed and order), frozen-append, LP and the gray outside-the-head reference; bars: std over seeds.
  The strip shows COCO mask mAP points lost on the 200-image protocol, where a run exists.}
  \label{fig:shots}
\end{figure*}

The F1 gain over fine-tuning is a low-shot one (Fig.~\ref{fig:shots}).
At ten shots no comparison with the mean FT order shows a gain for our methods, and its best order outperforms the \oursF{} on cars and aircraft.
At 20 shots (three seeds) the mean FT order outperforms the \oursF{} on pets and ties with it elsewhere, and the \ours{} outperforms FT on no set.
Below five shots (three seeds), the \ours{} outperforms FT's mean batch order on \lowAgentMeanWinsOne{}, \lowAgentMeanWinsTwo{} and \lowAgentMeanWinsThree{} of four sets at one, two and three shots, and the \oursF{} on \lowFormMeanWinsOne{}, \lowFormMeanWinsTwo{} and \lowFormMeanWinsThree{}; FT outperforms them in no cell, and they tie on aircraft throughout.
Against FT's best order most cells tie at three seeds: the \oursF{} outperforms it on \lowFormBestWinsTwo{} sets at two shots and the \ours{} on \lowAgentBestWinsThree{} at three.
At one shot no cross-validation fold exists, so every cross-validated method reuses its two-shot settings; against FT tuned on the test labels the \ours{} scores lower F1 on \lowOracleBelow{} of \lowOracleTested{} sets tested (one seed).
COCO loss$^\ast$ stays small at every shot count: \retAgentAllShots{} points on average over \nAgentRunsAll{} \ours{} runs, and \retFormAllShots{} over \nFormRunsAll{} \oursF{} runs.

\subsection{Forgetting and the joint-head trade-off}
\label{sec:res-forget}

\begin{table}[tb]
  \centering
  \small
\begingroup\setlength{\tabcolsep}{3pt}
\begin{adjustbox}{max width=\linewidth}
\begin{tabular}{lcccccc}
\toprule
 & \multicolumn{3}{c}{COCO points lost} & \multicolumn{3}{c}{Five shots, 90-row head} \\
\cmidrule(lr){2-4}\cmidrule(lr){5-7}
Method & 5 & 10 & 20 & parent AP & F1 alone & F1 joint \\
\midrule
Stock head & 0.0 & 0.0 & 0.0 & 56.0 & \na & \na \\
Linear probe (LP) & 42.7$^\ast$ & 49.0$^\ast$ & 43.1$^\ast$ & 23.4 & 21.3 & 16.4 \\
Fine-tuning (FT), mean of 3 orders & 49.1 & 51.3 & 51.3 & 5.8$^\S$ & 35.2 & 28.2 \\
\textcolor{gray}{FT, best order$^\ddagger$} & \multicolumn{3}{c}{\textcolor{gray}{as above}} & \textcolor{gray}{6.8$^\S$} & \textcolor{gray}{43.0} & \textcolor{gray}{34.3} \\
FT + LwF & 44.4$^\ast$ & 46.2$^\ast$ & 38.5$^\ast$ & \na & \na & \na \\
TFA-style, linear probe & 11.3 & \na & \na & \na & \na & \na \\
TFA-style, cosine head & 12.4 & \na & \na & \na & \na & \na \\
Frozen-append, best per set$^\ddagger$ & \textbf{0.1} & 0.5 & 0.8 & 63.5$^\S$ & 27.1 & 21.7 \\
Frozen-append, best in 90-row F1$^\ddagger$ & \textbf{0.1} & 1.2 & \textbf{0.5} & 63.7$^\S$ & 25.8 & 22.2 \\
WiSE-FT & 1.3 & \na & \na & \na & \na & \na \\
L2-SP & 37.3 & \na & \na & \na & \na & \na \\
SimpleShot + gate & 0.5$^\ast$ & \textbf{0.4}$^\ast$ & 0.8$^\ast$ & \textbf{43.2} & 39.1 & 33.4 \\
\ours{} (Opus) & 1.0 & 0.9 & \textbf{0.5} & 39.5 & 41.6 & 30.5 \\
\oursF{} & 1.0 & 1.1 & 1.3 & 43.0 & \textbf{47.2} & \textbf{42.0} \\
\bottomrule
\end{tabular}
\end{adjustbox}
\par\vspace{2pt}{\scriptsize Parent AP on 5000 val images (stock row: before the edit). $^\S$ Val images 201 to 400, where the stock head has 65.2; on them the \ours{} keeps 43.8 and the \oursF{} 52.2; read these rows against each other. Frozen-append COCO loss: images 201 to 400, best head picked at each shot count. WiSE-FT and L2-SP: five shots, each (set, seed, batch order) a run. $^\ddagger$ Best batch order per seed (FT) or best of 10 guarded variants per set (frozen-append; seven at 10 and 20 shots), picked on the test images. \na{}: not run.}
\endgroup
  \caption{\textbf{Forgetting and the 90-row head.}
  COCO mask mAP points lost at three shot counts and, at five shots, the AP of the task's COCO parent class after the edit and the new-class F1 with the new rows scored alone and inside the 90-row head (the 80 stock rows and the new rows in one softmax).
  Bold: best non-stock black row per column, $^\S$ cells excluded. Gray: FT at its best batch order ($^\ddagger$).
  TFA-style rows also train on COCO images (five shots only).
  $^\ast$ 200-image protocol; $^\S$ other COCO images, see the note.}
  \label{tab:forget}
\end{table}

Fine-tuning loses \retFtLo--\retFtHi{} points at five shots (Tab.~\ref{tab:forget}), LwF \retLwfLo--\retLwfHi{}.
The TFA-style baseline, which also trains on COCO images, loses \retTfaLo--\retTfaHi{} points, but its new-class F1 falls to \tfaNovelLo--\tfaNovelHi{} against \oursNovelLo--\oursNovelHi{} for our methods.
CD-ViTO forgets, from \cdvCocoStock{} to \cdvCocoFt{} box AP on COCO.

Retaining the 80-class mAP does not depend on the agent: nearest class mean and SimpleShot heads that append rows under the same guard and gate lose \retClassicLo--\retClassicHi{} points (Tab.~\ref{tab:program}).
The guard and the append-only structure retain the mean, whatever writes the rows.
Inside the head the \ours{} adds new-class accuracy at this retention.
Two standard fixes for fine-tuning do not reach that point: WiSE-FT retains COCO only at mixing weights where the new rows score \wiseScoreLo--\wiseScoreHi{} F1, and no L2-SP strength in our grid meets the guard, so it loses \lspDropLo--\lspDropHi{} points.

\noindent\textbf{Frozen-append baseline.}
Frozen-append (Sec.~\ref{sec:setup}) retains the 80-class mean as well as our methods, losing \faDropBest{} mAP points against \faDropAgent{} for the \ours{} and \faDropForm{} for the \oursF{} on val images 201 to 400.
It reaches a far higher parent-class AP: on the same images it goes from \faParentStock{} to \faParentFa{}, against \faParentAgent{} for the \ours{} and \faParentForm{} for the \oursF{}, and the cat AP from \faParentPetsStock{} to \faParentPetsFa{}, against \faParentPetsAgent{} and \faParentPetsForm{}.
Against its best variant per set, the \oursF{} outperforms it on \faFormWinsK{} of four sets by \faFormGainKLo{} to \faFormGainKHi{} F1 points and the \ours{} on \faAgentWinsK{} by \faAgentGainKLo{} to \faAgentGainKHi{} (two-level intervals above zero), and both tie with it on the rest; in the 90-row head they outperform it on \faFormWinsJ{} and \faAgentWinsJ{} sets.
TFA from the stock weights retains COCO only with new rows that reach \faTfaScore{} F1.
So at matched 80-class retention, gradient descent in the same head reaches lower new-class F1 than our methods.
Matched on the parent class instead, by one bias on each method's new rows under one guard on the parent-class AP, most of the gain disappears: the \oursF{} outperforms frozen-append on cars by \mpFormDiffCars{} joint F1 (\mpFormLoCars{} to \mpFormHiCars) and ties with it on \mpFormSame{} sets, and the \ours{} scores lower joint F1 than it on CUB and pets.

The 80-class mean does not show a loss that appears when the new rows and the 80 stock rows compete in one 90-row head.
There the new-class F1 at five shots falls for every method with both scores in Tab.~\ref{tab:forget}: the \ours{} goes from \jointAgentAlone{} to \jointAgentJoint{}, the \oursF{} from \jointFormAlone{} to \jointFormJoint{}, and FT at its best order from \jointCfBestAlone{} to \jointCfBestJoint{}.
The \oursF{} outperforms FT's best order in joint F1 on cars (\jointFormCars{} against \jointCfBestCars{}), CUB (\jointFormCub{} against \jointCfBestCub{}) and pets (\jointFormPets{} against \jointCfBestPets{}) and ties with it on aircraft (\jointFormAircraft{} against \jointCfBestAircraft{}), but FT retains almost none of the parent class (\jointCfParentStock{} to \jointCfParentAfter{}).
These are point estimates; for the \oursF{} tuned in the joint head, the two-level interval over FT's best order excludes zero on CUB only.
The COCO bird AP on 5000 val images goes from \jointParentCubStock{} to \jointParentCubAgent{} (\ours{}) and \jointParentCubForm{} (\oursF{}), cat from \jointParentPetsStock{} to \jointParentPetsAgent{} and \jointParentPetsForm{}, and airplane from \jointParentAircraftStock{} to \jointParentAircraftAgent{} and \jointParentAircraftForm{}, while car barely moves (\jointParentCarsStock{} to \jointParentCarsAgent{} and \jointParentCarsForm{}).
Crediting a new row's detection to the parent class does not remove the drop: under that rule the parent AP is lower still in \jointHierWorseForm{} of \jointHierRunsForm{} \oursF{} runs.

Tuning in the joint head does not fix the loss.
A scalar subtracted from the new-row biases, picked on separate COCO images to keep the parent class within two AP points, lowers joint F1 by \jcalCostLo{} to \jcalCostHi{} and the parent still loses \jcalDropLo{} to \jcalDropHi{} AP points; re-tuning the \oursF{} on joint F1 meets the parent guard in \jcalTunedPassed{} of \jcalTunedRuns{} cells.
The parent class and the new classes trade against each other.
The average does not show a tail: in the worst run (\ours{}, CUB, seed~0) COCO mAP drops \retWorst{} points because its new rows outscore stock rows on look-alike COCO objects like zebra and sheep, but the guard score the agent reads never exceeds \guardMax{} points in \nAgentRunsGuard{} runs.

\subsection{Program analysis}
\label{sec:res-program}
\label{sec:analysis}

\begin{table*}[t]
  \centering
  \small
\begin{minipage}[t]{0.325\linewidth}
\begingroup\setlength{\tabcolsep}{3pt}
\begin{adjustbox}{max width=\linewidth}
\begin{tabular}{lcccc}
\toprule
\multicolumn{5}{l}{\emph{(a) Heads with the gate of Eq.~\ref{eq:formula}; parts removed}} \\
\midrule
 & \multicolumn{3}{c}{F1 by shots} & COCO \\
Head & 5 & 10 & 20 & lost \\
\midrule
\oursF{} & \textbf{48.3} & \textbf{55.7} & \textbf{62.5} & 0.9 \\
Nearest mean + gate & 9.7 & 14.0 & 15.6 & \textbf{0.1} \\
SimpleShot + gate & 39.1 & 43.6 & 47.9 & 0.5 \\
LDA + gate & 24.2 & 26.5 & 25.7 & 0.5 \\
\oursF{} without the gate & 13.6 & 15.1 & 15.0 & 22.7 \\
\oursF{} without nuisance removal & 35.2 & 41.3 & 46.9 & 0.4 \\
\oursF{} without the class term & 0.0 & 0.0 & 0.0 & 0.7 \\
\bottomrule
\end{tabular}
\end{adjustbox}
\par\vspace{2pt}{\scriptsize F1 $\times$100, mean over four sets and seeds 0--2. COCO lost: 200 images, five shots.}
\endgroup
\end{minipage}
\hfill
\begin{minipage}[t]{0.31\linewidth}
\begingroup\setlength{\tabcolsep}{3pt}
\begin{adjustbox}{max width=\linewidth}
\begin{tabular}{lcc}
\toprule
\multicolumn{3}{l}{\emph{(b) Replay without the agent}} \\
\midrule
Cars recipe, seed 0, 20 shots & Replay & Live \\
\midrule
Cars, seeds 1 and 2, 20 shots & 87.9 & 92.8 \\
Cars, seed 0, 5 shots & 72.4 & 61.6 \\
Cars, seed 0, 10 shots & 83.4 & 81.8 \\
CUB, seed 0, 20 shots & 52.4 & 55.8 \\
Pets, seed 0, 20 shots & 68.7 & 76.6 \\
Aircraft, seed 0, 20 shots & 13.6 & 30.2 \\
\bottomrule
\end{tabular}
\end{adjustbox}
\endgroup
\end{minipage}
\hfill
\begin{minipage}[t]{0.325\linewidth}
\begingroup\setlength{\tabcolsep}{3pt}
\begin{adjustbox}{max width=\linewidth}
\begin{tabular}{lcc}
\toprule
\multicolumn{3}{l}{\emph{(c) Terms of the formula in the agent's programs}} \\
\midrule
Term & Independent (243) & After formula (195) \\
\midrule
Class-mean rows & 242 (100\%) & 194 (99\%) \\
Removal of shared directions & 225 (93\%) & 182 (93\%) \\
Explicit gate & 182 (75\%) & 136 (70\%) \\
All three terms & 166 (68\%) & 124 (64\%) \\
Exact formula shape & 45 (19\%) & 24 (12\%) \\
\bottomrule
\end{tabular}
\end{adjustbox}
\par\vspace{2pt}{\scriptsize Programs counted by static rules, hand-checked on a sample. ``After formula'': written once the formula was fixed. Warm-start runs are left out.}
\endgroup
\end{minipage}
  \caption{\textbf{The program.}
  \emph{(a)} The \oursF{} against closed-form heads with the same gate and guard, and with one part removed.
  \emph{(b)} Replay without the agent.
  \emph{(c)} Share of the agent's programs that contain each term.
  Bold in (a): best per column (lowest COCO loss).}
  \label{tab:program}
\end{table*}

\texttt{recipe.py} replays without the agent (Tab.~\ref{tab:program}b).
The program of one 20-shot cars run, applied to two other seeds, scores \replayCarsOne{} and \replayCarsTwo{} F1, against \liveCarsOne{} and \liveCarsTwo{} for the live agent; moved to the other three sets, a program scores \replayXsetLo--\replayXsetHi{} F1 on average where the live agent scores \liveXsetLo--\liveXsetHi{}, so replay across sets is partial.

The programs contain a gate, class means and removal of shared directions: \convAll{} of \convTotal{} independent programs contain all three, and \convAfter{} of \convAfterTotal{} written after the formula was fixed (Tab.~\ref{tab:program}c).
However, the exact formula shape is rare (\convExact{} of \convTotal), and the prompt names the allowed kinds of edit (class means, directions, thresholds), so part of the overlap comes from the prompt.
The \oursF{} fixes this form as code: a nearest class mean classifier with a gate and that removal.
Nearest class mean, SimpleShot and shrinkage LDA heads, each given the same gate and guard, score lower F1 than it on \nClassicLose{} of 12 set-by-shot cells.
Removing the gate lowers five-shot F1 on every set, and omitting that removal lowers it on cars, CUB and pets.
On a second, disjoint draw of ten classes per set, the unchanged \oursF{} ties with or outperforms FT on \freshTieOrBeat{} of four sets and outperforms LP on \freshBeatCtwo{}; on ten sparrows it scores lower F1 than FT's best order (supplement F).

The prompt bans closed-form regression and least-squares fits of weights to labels and allows simple statistics.
The audit flags \noldaFlagged{} programs with an LDA-style metric, \noldaConvergeFlagged{} of them independent programs (Sec.~\ref{sec:method-harness}).
Among them is no main five-shot run, so all \noldaFiveCells{} five-shot comparisons stay unchanged; at ten shots \noldaHeadlineTen{} main runs are excluded and the gap to FT's best order moves by at most \noldaTenMove{} points (supplement~D).
A scan of the logs finds a least-squares, pseudo-inverse, SVD or eigen-decomposition call in \auditRunsFit{} of \auditRunsTotal{} audited runs, mostly for allowed statistics.

The edit does not need the detector's class head to store the classes.
Given the backbone features of the stock detector's boxes, the agent edits a new linear head with F1 close to that of nearest class mean, higher on \featNcmBeat{} of \featSets{} sets and lower on \featNcmLose{} (two-level intervals, three seeds).
On these features the linear head has a higher F1 than the \ours{} on \featOverRows{} sets, pets and ships, and a lower F1 on \featUnderRows{}.
The agent recovers the standard nearest class mean classifier from five images per class without a gradient, and does not exceed it.

\noindent\textbf{Decomposition of the gain.}
We decompose the change inside the class rows into a shared part, which raises the presence logit (the gate), and a class part, which adds knowledge of the classes (supplement H).
On the fine-grained sets the shared part finds the object but cannot assign its class (macro-F1 \gateMacroLo--\gateMacroHi), the class part alone scores far below the full edit, and a random edit of the same size scores at most \randomEditMax{}, so both parts are needed.
On SAM~3 nuclei the gate gives most of the gain: the stock presence logit puts every score below the threshold, and the edit raises F1 from zero to \samNucleiLo--\samNucleiHi{} at 20 shots, where Adam~\cite{adam} on the same tensors reaches \samNucleiAdam{} and a per-project trained head \samNucleiTrained.

\subsection{Effect of prior knowledge}
\label{sec:res-prior}

\begin{figure}[tb]
  \centering
  \includegraphics[width=\linewidth]{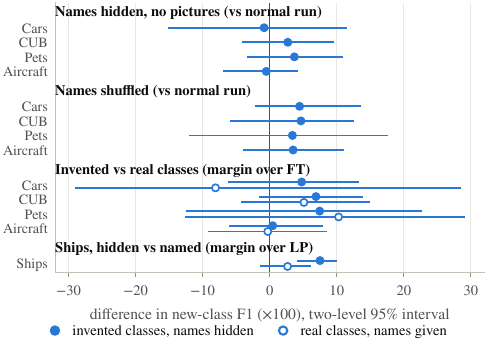}
  \caption{\textbf{Tests of the agent's use of class names at five shots.}
  One block per test, one row per set: names hidden, names shuffled, invented classes, and satellite ships.
  Dots show the change in the agent's F1 against its unmodified run (hidden and shuffled names), or its margin over FT, one batch order (invented against real classes), or over LP (ships; hidden names filled, names given hollow).
  Bars: two-level 95\% intervals (seeds, then test images; \bootTwoLevelResamples{} resamples), three seeds.}
  \label{fig:prior}
\end{figure}

To test whether the agent writes the edit from what it learned in pretraining about cars, birds and cats, we remove that source four ways (Fig.~\ref{fig:prior}).
The intervals come from three seeds and are wide, so a test with no detectable change says we cannot tell.
Hiding the class names and the pictures changes the agent's F1 by at most \priorStripMax{} points on any set at five shots, and shuffling the names moves it by \priorPermLo{} to \priorPermHi{} points; \priorHidShufExcl{} of the \priorHidShufTotal{} intervals exclude zero.
On invented classes (real classes paired into groups that do not exist), the agent's F1 margin over FT runs from \priorMixLo{} to \priorMixHi{} points over the four sets, and \priorMixExcl{} of the four intervals exclude zero.
A static audit flags \nRecipesFlagged{} of \nRecipesAudited{} programs for a class-name word or fixed class ids, and none is the best run of its cell.

Ships test this where the prior is weak: a vision-language model classifies \vlmShipsAcc{} of vessel crops correctly zero-shot, against \vlmCarsAcc{} on cars.
The agent with hidden names outperforms LP by \shipsHidProbe{} F1 points (\shipsHidProbeLo--\shipsHidProbeHi) and FT's mean order by \shipsHidMean{} (\shipsHidMeanLo--\shipsHidMeanHi), and ties with FT's best order.
The agent with the names ties with FT (\statShipsNamed{} points to the mean order; interval spans zero), and hiding the names raises its F1 (\shipsHidNamed{} points, \shipsHidNamedLo--\shipsHidNamedHi, in \shipsHidNamedWon{} of \shipsSeeds{} seeds).
\section{Limitations}
\label{sec:limits}

The outside-the-head classifier scores \tsScoreLo--\tsScoreHi{} F1 on the fine-grained sets at five shots and outperforms the \ours{} and the \oursF{} on \tsAbove{} of \tsSets{} sets.
Our gain over full-decoder fine-tuning disappears from ten shots, where fine-tuning ties with or outperforms both in F1 (Sec.~\ref{sec:res-shots}).
CD-ViTO, a published detector with another backbone, exceeds both in mean F1 on \cdvWinsPub{} of \cdvSetsTotal{} sets, and on the CD-FSOD benchmark our edits score a lower box AP than the best published results, so we make no benchmark claim.
In its worst run the \ours{} lowers COCO mAP by \retWorst{} points, and in the 90-row head the parent-class AP drops by up to \jointDropHi{} points for both methods.
Frozen-append reaches a higher parent-class AP than both, but when the parent-class AP is matched it ties with the \oursF{} in joint F1 on \mpFormSame{} of four sets (Sec.~\ref{sec:res-forget}).
The \ours{} depends on the agent model: Sonnet~5 fails in the same harness, with F1 of \sonnetLo--\sonnetHi{} on all \nSonnetCells{} cells against Opus's \opusCellsLo--\opusCellsHi.
Across four Opus runs the standard deviation of F1 is \repeatSdCars{} on cars and \repeatSdPets{} on pets, and an agent run costs \$\agentUsdLo--\$\agentUsdHi{}.
On the fine-grained sets, a vision-language model classifies \vlmCropLo--\vlmCropHi\% of ground-truth crops correctly, against \headNameLo--\headNameHi\% for our two methods and fine-tuning on the objects each finds.
On CLIP, over datasets and shot counts, the \ours{} improves on zero-shot accuracy in most cases but scores below the best baseline given the same labels in nearly all cases (supplement~I).
Absolute F1 is low on aircraft and ships (Tab.~\ref{tab:main}).
\section{Conclusion}
\label{sec:conclusion}

We asked whether a coding agent can replace gradient descent when a frozen vision model's head must learn new classes from a few images.
Partly: at five shots, inside the same head, the \ours{} outperforms fine-tuning in F1 on \nWinAgent{} of four fine-grained sets and ties on the rest, and the gain disappears by ten shots.
It keeps the mean COCO mAP over the 80 classes but lowers the parent-class AP in a shared head, and once that class is matched, gradient descent with the old rows frozen ties with or outperforms our edits except on cars.
The programs, a nearest class mean classifier with a gate and removal of shared directions, replay without the agent, and we find no sign that prior knowledge of the classes explains the result.
A classifier outside the head reaches a higher F1 on every set.
We release the harness and the \oursF{} at \url{https://github.com/Khayrulbuet13/agent-edit}.

\section*{LLM Usage Statement}
\label{sec:llm}
This work evaluates LLM coding agents, so LLMs are part of the method: Claude Opus~5.5, GPT~6.1 Sol and Claude Sonnet~5 wrote the weight-editing programs that we study, inside the harness of Sec.~\ref{sec:method}.
Beyond the method, we used Claude Code (Anthropic) to assist with writing and editing the code, to run the experiments that the authors designed, to edit the wording and tone of the text, and to check the reference list and the manuscript for errors.
It did not generate research ideas, hypotheses, claims or conclusions.
The authors take full responsibility for all submitted content.

{
    \small
    \bibliographystyle{ieeenat_fullname}
    \bibliography{refs}
}

\clearpage
\maketitlesupplementary
\setcounter{section}{0}
\setcounter{table}{0}
\setcounter{figure}{0}
\setcounter{equation}{0}
\renewcommand{\thesection}{\Alph{section}}
\renewcommand{\thetable}{S\arabic{table}}
\renewcommand{\thefigure}{S\arabic{figure}}
\renewcommand{\theequation}{S\arabic{equation}}
\renewcommand{\theHsection}{supp.\Alph{section}}
\renewcommand{\theHtable}{supp.\arabic{table}}
\renewcommand{\theHfigure}{supp.\arabic{figure}}
\renewcommand{\theHequation}{supp.\arabic{equation}}

In this supplementary material, we provide the full results (Sec.~\ref{sec:supp-A}), the baseline protocol (Sec.~\ref{sec:supp-B}), the harness and the task prompt (Sec.~\ref{sec:supp-C}), the fitting audit (Sec.~\ref{sec:supp-D}), ablations and controls (Sec.~\ref{sec:supp-E}), the agent's programs (Sec.~\ref{sec:supp-F}), qualitative examples (Sec.~\ref{sec:supp-G}), the SAM~3 study (Sec.~\ref{sec:supp-H}), the CLIP study (Sec.~\ref{sec:supp-I}), and the CD-ViTO and CD-FSOD setup (Sec.~\ref{sec:supp-J}).
All numbers come from the same result files as the main paper.

\section{Full results}
\label{sec:supp-A}

\subsection{F1 on every set and shot count}
\label{sec:supp-A-pmt}

Besides the linear probe (LP), the cosine head and full-decoder fine-tuning (FT) of \Msec{sec:setup}, Tab.~\ref{tab:supp-pmt-fg} reports two training-free heads: Prototype, the centered class mean of the labeled queries, and Tip-Adapter~\cite{zhang2022tipadapter}, which adds its cache term on top of the Prototype rows.
The two FT rows are the mean and the best of three batch orders per seed (Sec.~\ref{sec:supp-B-orders}).

\begin{table*}[!t]
  \centering
  \scriptsize
\begingroup\setlength{\tabcolsep}{2pt}
\begin{adjustbox}{max width=\linewidth}
\begin{tabular}{lcccccccccccc}
\toprule
 & \multicolumn{6}{c}{Cars} & \multicolumn{6}{c}{CUB} \\
\cmidrule(lr){2-7}\cmidrule(lr){8-13}
Method ($N$) & 1 & 2 & 3 & 5 & 10 & 20 & 1 & 2 & 3 & 5 & 10 & 20 \\
\midrule
Prototype & 14.1\pmstd{1.1}$^\star$ & 13.2\pmstd{2.3} & 10.4\pmstd{6.9} & 18.4\pmstd{3.0} & 14.4\pmstd{5.4} & 21.4\pmstd{4.1} & 11.6\pmstd{0.7}$^\star$ & 6.9\pmstd{2.2} & 14.3\pmstd{0.4} & 12.0\pmstd{3.2} & 18.9\pmstd{2.4} & 19.5\pmstd{2.8} \\
Linear probe (LP) & 13.6\pmstd{3.9}$^\star$ & 20.4\pmstd{4.6} & 26.1\pmstd{4.3} & 30.6\pmstd{4.0} & 49.3\pmstd{6.4} & 61.5\pmstd{2.4} & 13.0\pmstd{1.0}$^\star$ & 12.8\pmstd{2.1} & 15.5\pmstd{2.8} & 15.2\pmstd{3.7} & 17.9\pmstd{2.5} & 23.5\pmstd{1.6} \\
Tip-Adapter & 11.2\pmstd{2.7}$^\star$ & 8.6\pmstd{7.2} & 11.8\pmstd{7.9} & 16.8\pmstd{1.8} & 14.2\pmstd{4.9} & 18.3\pmstd{3.9} & 9.9\pmstd{4.5}$^\star$ & 4.1\pmstd{3.4} & 5.9\pmstd{5.5} & 10.3\pmstd{4.0} & 20.4\pmstd{2.3} & 17.5\pmstd{1.5} \\
Cosine head & 13.4\pmstd{3.7}$^\star$ & 19.5\pmstd{6.3} & 22.4\pmstd{6.8} & 33.0\pmstd{4.5} & 49.6\pmstd{7.5} & 64.2\pmstd{1.7} & 11.2\pmstd{2.7}$^\star$ & 10.2\pmstd{6.8} & 13.6\pmstd{6.4} & 19.2\pmstd{3.8} & 19.0\pmstd{2.1} & 23.2\pmstd{0.6} \\
Fine-tuning (FT), mean of 3 orders & 11.0\pmstd{3.7}$^\star$ & 20.5\pmstd{6.4} & 32.8\pmstd{6.6} & 58.1\pmstd{13.4} & 80.8\pmstd{14.7} & 90.3\pmstd{9.7} & 10.3\pmstd{2.3}$^\star$ & 10.7\pmstd{1.6} & 15.3\pmstd{2.0} & 22.5\pmstd{6.9} & 46.1\pmstd{6.6} & \textbf{58.2}\pmstd{4.1} \\
\textcolor{gray}{FT, best order$^\ddagger$} & \textcolor{gray}{14.3\pmstd{4.5}$^\star$} & \textcolor{gray}{29.7\pmstd{10.7}} & \textcolor{gray}{38.2\pmstd{5.1}} & \textcolor{gray}{71.2\pmstd{11.5}} & \textcolor{gray}{96.4\pmstd{2.2}} & \textcolor{gray}{97.9\pmstd{0.3}} & \textcolor{gray}{15.6\pmstd{2.0}$^\star$} & \textcolor{gray}{13.2\pmstd{1.6}} & \textcolor{gray}{18.8\pmstd{3.6}} & \textcolor{gray}{28.4\pmstd{7.4}} & \textcolor{gray}{50.3\pmstd{8.7}} & \textcolor{gray}{63.7\pmstd{4.8}} \\
\oursF{} & 22.2\pmstd{6.2}$^\star$ & 31.9\pmstd{5.0} & \textbf{44.2}\pmstd{5.2} & \textbf{65.0}\pmstd{5.6} & 81.5\pmstd{3.3} & 90.4\pmstd{1.9} & \textbf{24.4}\pmstd{7.0}$^\star$ & \textbf{27.5}\pmstd{3.7} & \textbf{31.7}\pmstd{8.8} & \textbf{39.6}\pmstd{3.7} & 45.0\pmstd{3.9} & 52.1\pmstd{2.7} \\
\ours{} (Opus) & \textbf{22.8}\pmstd{7.0} & \textbf{37.1}\pmstd{7.1} & 43.1\pmstd{6.0} & 58.1\pmstd{12.6} & \textbf{81.7}\pmstd{4.4} & \textbf{92.5}\pmstd{0.5} & 18.3\pmstd{2.1} & 22.8\pmstd{9.7} & 30.3\pmstd{7.0} & 32.4\pmstd{5.1} & \textbf{46.5}\pmstd{8.4} & 51.5\pmstd{4.4} \\
\midrule
 & \multicolumn{6}{c}{Pets} & \multicolumn{6}{c}{Aircraft} \\
\cmidrule(lr){2-7}\cmidrule(lr){8-13}
Method ($N$) & 1 & 2 & 3 & 5 & 10 & 20 & 1 & 2 & 3 & 5 & 10 & 20 \\
\midrule
Prototype & 6.3\pmstd{2.1}$^\star$ & 9.5\pmstd{3.6} & 14.5\pmstd{0.6} & 17.1\pmstd{0.5} & 21.4\pmstd{2.3} & 23.2\pmstd{2.1} & 13.9\pmstd{4.5}$^\star$ & 11.0\pmstd{3.8} & 13.5\pmstd{1.2} & 10.3\pmstd{2.3} & 15.4\pmstd{0.9} & 16.7\pmstd{1.6} \\
Linear probe (LP) & 5.9\pmstd{2.0}$^\star$ & 9.5\pmstd{2.2} & 12.9\pmstd{2.4} & 17.4\pmstd{3.6} & 26.8\pmstd{4.5} & 33.1\pmstd{4.0} & 14.5\pmstd{1.3}$^\star$ & 14.1\pmstd{2.5} & \textbf{19.1}\pmstd{4.3} & 22.1\pmstd{6.3} & 31.1\pmstd{5.6} & 33.6\pmstd{5.1} \\
Tip-Adapter & 9.1\pmstd{2.6}$^\star$ & 7.6\pmstd{4.3} & 13.4\pmstd{0.6} & 16.7\pmstd{1.0} & 20.4\pmstd{3.3} & 19.1\pmstd{2.6} & 13.8\pmstd{4.5}$^\star$ & 6.6\pmstd{5.8} & 11.6\pmstd{1.2} & 9.9\pmstd{2.1} & 14.3\pmstd{1.4} & 15.0\pmstd{1.3} \\
Cosine head & 3.9\pmstd{0.5}$^\star$ & 9.2\pmstd{4.7} & 11.4\pmstd{3.1} & 14.8\pmstd{1.7} & 25.3\pmstd{2.5} & 30.5\pmstd{3.7} & \textbf{17.6}\pmstd{0.9}$^\star$ & 15.1\pmstd{0.6} & 15.2\pmstd{3.4} & 20.2\pmstd{2.9} & 29.8\pmstd{1.2} & 34.0\pmstd{4.8} \\
Fine-tuning (FT), mean of 3 orders & 12.1\pmstd{0.6}$^\star$ & 12.5\pmstd{2.5} & 23.1\pmstd{7.6} & 39.4\pmstd{3.0} & 65.6\pmstd{7.7} & \textbf{85.0}\pmstd{4.3} & 16.5\pmstd{6.8}$^\star$ & 11.4\pmstd{1.3} & 18.6\pmstd{2.5} & 20.9\pmstd{2.3} & \textbf{32.8}\pmstd{4.0} & \textbf{37.5}\pmstd{11.5} \\
\textcolor{gray}{FT, best order$^\ddagger$} & \textcolor{gray}{15.3\pmstd{0.5}$^\star$} & \textcolor{gray}{15.6\pmstd{3.4}} & \textcolor{gray}{32.1\pmstd{11.9}} & \textcolor{gray}{48.6\pmstd{7.4}} & \textcolor{gray}{76.1\pmstd{5.9}} & \textcolor{gray}{90.4\pmstd{3.6}} & \textcolor{gray}{17.9\pmstd{6.1}$^\star$} & \textcolor{gray}{15.9\pmstd{3.7}} & \textcolor{gray}{21.8\pmstd{2.8}} & \textcolor{gray}{23.9\pmstd{1.7}} & \textcolor{gray}{38.1\pmstd{4.7}} & \textcolor{gray}{42.6\pmstd{11.7}} \\
\oursF{} & 12.4\pmstd{4.0}$^\star$ & \textbf{39.6}\pmstd{4.6} & \textbf{46.8}\pmstd{5.9} & \textbf{61.7}\pmstd{4.1} & \textbf{71.2}\pmstd{5.0} & 74.7\pmstd{2.6} & 11.1\pmstd{3.4}$^\star$ & \textbf{16.3}\pmstd{0.6} & 17.1\pmstd{1.4} & 22.3\pmstd{2.9} & 25.9\pmstd{4.3} & 32.8\pmstd{6.1} \\
\ours{} (Opus) & \textbf{23.0}\pmstd{7.8} & 31.5\pmstd{11.5} & 46.7\pmstd{5.2} & 52.3\pmstd{11.2} & 64.9\pmstd{3.9} & 78.8\pmstd{2.8} & 15.5\pmstd{2.3} & 15.6\pmstd{1.7} & 16.5\pmstd{2.8} & \textbf{23.6}\pmstd{4.3} & 25.8\pmstd{8.5} & 32.6\pmstd{2.2} \\
\bottomrule
\end{tabular}
\end{adjustbox}
\par\vspace{2pt}{\scriptsize Held-out instance macro-F1 ($\times$100) at IoU 0.5; $N$ train images per class. With 1 to 3 shots the cross-validation folds hold out one shot per class, so $N{=}2$ has two folds, $N{=}3$ three and $N{=}1$ none; $^\star$ at one shot each of these methods reuses the setting it picked at two shots, and the agent runs unchanged. Mean$\pm$std over seeds; bold: best of each column. $^\ddagger$ best of three batch orders per seed, picked on the test labels (gray, not in the bold comparison).}
\endgroup
  \caption{\textbf{Instance macro-F1 on the four fine-grained sets.}
  Box IoU for cars, CUB and aircraft, mask IoU for pets; five seeds at 5 and 10 shots (three for Prototype, Tip-Adapter and the cosine head) and three below five shots and at 20 shots.
  The baselines were tuned by cross-validation on the training images, with five folds from five shots on (Sec.~\ref{sec:supp-B}).
  Some two-shot Tip-Adapter runs score zero.
  Its cache term grows with the number of training images and its setting is picked on half of them, so after the refit on all of them the no-object row wins every query.
  Ships, at five shots, are in \Mtab{tab:main}.}
  \label{tab:supp-pmt-fg}
\end{table*}

\subsection{AP, AP50 and top-1 localization}
\label{sec:supp-A-ap}

Tab.~\ref{tab:supp-ap} gives AP, AP50 and top-1 localization at five shots for the main rows of \Mtab{tab:main}.
One scorer handles every method: its 100 best detections per image, with no score threshold, are scored as box AP over IoU 0.5 to 0.95 (AP) and at IoU~0.5 (AP50).
Top-1 localization takes the highest-scoring detection of each test image and counts it when its box has an IoU of at least 0.5 with a true box of its class.

\newcommand{\suppApCells}[1]{\mval{#1Ap} & \mval{#1Apfifty} & \mval{#1ToponeLoc}}
\newcommand{\suppApCellsGray}[1]{\textcolor{gray}{\mval{#1Ap}} & \textcolor{gray}{\mval{#1Apfifty}} & \textcolor{gray}{\mval{#1ToponeLoc}}}

\begin{table*}[t]
  \centering
  \footnotesize
  \setlength{\tabcolsep}{4pt}
  \begin{tabular}{lcccccccccccc}
    \toprule
    & \multicolumn{3}{c}{Cars} & \multicolumn{3}{c}{CUB} & \multicolumn{3}{c}{Pets} & \multicolumn{3}{c}{Aircraft} \\
    \cmidrule(lr){2-4}\cmidrule(lr){5-7}\cmidrule(lr){8-10}\cmidrule(lr){11-13}
    Method & AP & AP50 & Top-1 & AP & AP50 & Top-1 & AP & AP50 & Top-1 & AP & AP50 & Top-1 \\
    \midrule
    Stock head & \suppApCells{stockTagApAllCarsNfive} & \suppApCells{stockTagApAllCubNfive} & \suppApCells{stockTagApAllPetsNfive} & \suppApCells{stockTagApAllAircraftNfive} \\
    Linear probe (LP) & \suppApCells{ctwoTagApAllCarsNfive} & \suppApCells{ctwoTagApAllCubNfive} & \suppApCells{ctwoTagApAllPetsNfive} & \suppApCells{ctwoTagApAllAircraftNfive} \\
    Cosine head & \suppApCells{cfiverTagApAllCarsNfive} & \suppApCells{cfiverTagApAllCubNfive} & \suppApCells{cfiverTagApAllPetsNfive} & \suppApCells{cfiverTagApAllAircraftNfive} \\
    Fine-tuning (FT), mean of 3 orders & \suppApCells{cfivefTagApAllCarsNfive} & \suppApCells{cfivefTagApAllCubNfive} & \suppApCells{cfivefTagApAllPetsNfive} & \suppApCells{cfivefTagApAllAircraftNfive} \\
    \textcolor{gray}{FT, best order$^\ddagger$} & \suppApCellsGray{cfivefTagApAllBestCarsNfive} & \suppApCellsGray{cfivefTagApAllBestCubNfive} & \suppApCellsGray{cfivefTagApAllBestPetsNfive} & \suppApCellsGray{cfivefTagApAllBestAircraftNfive} \\
    Frozen-append, best per set$^\ddagger$ & \suppApCells{faBestTagApAllCarsNfive} & \suppApCells{faBestTagApAllCubNfive} & \suppApCells{faBestTagApAllPetsNfive} & \suppApCells{faBestTagApAllAircraftNfive} \\
    CD-ViTO & \suppApCells{cdvitoTagApAllCarsNfive} & \suppApCells{cdvitoTagApAllCubNfive} & \suppApCells{cdvitoTagApAllPetsNfive} & \suppApCells{cdvitoTagApAllAircraftNfive} \\
    SimpleShot + gate & \suppApCells{simpleshotGateTagApAllCarsNfive} & \suppApCells{simpleshotGateTagApAllCubNfive} & \suppApCells{simpleshotGateTagApAllPetsNfive} & \suppApCells{simpleshotGateTagApAllAircraftNfive} \\
    \textcolor{gray}{Two-stage (stock boxes, frozen classifier)$^\ddagger$} & \suppApCellsGray{twoStageTagApAllCarsNfive} & \suppApCellsGray{twoStageTagApAllCubNfive} & \suppApCellsGray{twoStageTagApAllPetsNfive} & \suppApCellsGray{twoStageTagApAllAircraftNfive} \\
    \midrule
    \ours{} (Opus) & \suppApCells{opusTagApAllOvlCarsNfive} & \suppApCells{opusTagApAllOvlParCubNfive} & \suppApCells{opusTagApAllOvlParPetsNfive} & \suppApCells{opusTagApAllOvlAircraftNfive} \\
    \oursF{} & \suppApCells{formulaGuardTagApAllCarsNfive} & \suppApCells{formulaGuardTagApAllCubNfive} & \suppApCells{formulaGuardTagApAllPetsNfive} & \suppApCells{formulaGuardTagApAllAircraftNfive} \\
    \bottomrule
  \end{tabular}
  \caption{\textbf{AP, AP50 and top-1 localization at five shots.}
  Box AP over IoU 0.5 to 0.95, box AP at IoU~0.5 and top-1 localization, all multiplied by 100, on the test images of each fine-grained set.
  Pets uses box AP for every method, because CD-ViTO and the two-stage reference return no masks.
  Mean over five seeds, and over three for the cosine head and SimpleShot.
  Gray rows are the references of \Mtab{tab:main}; $^\ddagger$ marks rows selected using the test labels.}
  \label{tab:supp-ap}
\end{table*}

\subsection{Statistics}
\label{sec:supp-A-stats}

In the two-level bootstrap (\Msec{sec:setup}), the seeds share the \nEvalFine{} test images of a set, so the images are one source of variation and the seeds are another.
A one-level bootstrap over test images alone pools the test images of all five seeds as if they were separate images, so it counts the same photo once per seed.
Its intervals are \ciWidthRatio{} times narrower in the median, so we report the wider two-level intervals.

\section{Baseline protocol}
\label{sec:supp-B}

\subsection{Selection and grids}
\label{sec:supp-B-cv}

Each method has one fixed grid (Tab.~\ref{tab:supp-grids}).
We pick the grid point by stratified 5-fold cross-validation on the training images alone, refit on all training images with the winner, and score the held-out images once.
For a grid point, we fit on four folds, predict the fifth, add up true positives, false positives and missed instances over the five held-out folds, and take the macro-F1 of the sums.
Below five shots, each fold holds out one shot per class, so two shots give two folds and three shots give three; at one shot, each method reuses the setting it picked at two shots.
The training-free heads (Prototype, Tip-Adapter and the \oursF{}) have no learned no-object row, so their fit ends with a sweep of one bias offset on the class rows (the offset $c$ of the \oursF{}) on the F1 of the fitting images, which runs inside every fold.
We widened the FT grid until its optimum lay inside it: a narrower grid, with learning rates up to $3\cdot10^{-4}$ and up to 12 epochs, put the optimum on the upper edge for every set.

\begin{table}[t]
  \centering
  \scriptsize
  \setlength{\tabcolsep}{3pt}
  \begin{tabular}{@{}p{0.17\linewidth}p{0.8\linewidth}@{}}
    \toprule
    Method & Grid \\
    \midrule
    Prototype & Row scale, as a multiple of the stock row norm, in $\{1,2,4,\dots,128\}$, and two target rules (all queries with IoU at least 0.5, or one query per object). \\
    LP & L-BFGS, cross-entropy over the $K$ classes and the no-object row, no-object weight 0.1, one query per object as target: weight decay 0 and 15 log-spaced values from $10^{-8}$ to $10^{-2}$. \\
    Tip-Adapter & On top of the Prototype rows: $\alpha\in\{0.5,1,2,5\}$ and $\beta\in\{1,3,5.5,10\}$. \\
    Cosine head & Imprinted initial rows, full-batch Adam, no-object weight 0.1: scale $\{20,50,100,200\}$, learning rate $\{10^{-2},3\cdot10^{-2}\}$, steps $\{500,2000\}$. \\
    FT & AdamW (weight decay 0.05) on the whole decoder, started from the stock weights plus the LP rows: learning rate $\{10^{-5},2.2\cdot10^{-5},4.6\cdot10^{-5},10^{-4},2.2\cdot10^{-4},4.6\cdot10^{-4},10^{-3}\}$ and epochs $\{1,2,4,8,16,24\}$. \\
    LwF & FT with a distillation term of weight 0.01. \\
    WiSE-FT & The FT weights mixed with the stock weights: mixing weight from 0 to 0.3 in steps of 0.005 and from 0.35 to 1 in steps of 0.05; we keep the largest one whose drop on the 20 guard images is within one point. \\
    L2-SP & FT retrained at its selected learning rate and epochs with a penalty $\lambda\lVert\theta-\theta_{\mathrm{stock}}\rVert^2$ on all tensors, $\lambda\in\{10^{-4},10^{-3},10^{-2},0.1,1,10\}$; we keep the smallest one that meets the guard, or the one with the smallest drop on the guard images if none does. \\
    TFA-style & LP and the cosine head as a joint head over the 80 stock rows, the new rows and the no-object row, trained on the training images of the set and five COCO training images per class (five shots only); the grid point is picked by a combined score of new-class F1 and base-class retention on separate COCO training images. \\
    \oursF{} & Gate gain $\{0,2,4,8,16,32\}$, class gain $\{1,2,4,8,16,32\}$, removed directions $\{0,5,10,20,30,40,50\}$ (capped at a third of the number of primary queries minus the number of classes), the base row (COCO parent row or no-object row) for cars, aircraft and ships, and the bias offset swept from $-80$ to $80$ in steps of 5 on the fit F1. \\
    Closed form & Nearest class mean, SimpleShot and shrinkage LDA, fitted on the same primary queries as the \oursF{} and given its gate: the row scale, and for LDA the shrinkage $\{0.9,0.95,0.99,0.999\}$ times scale $\{8,16,32,64\}$. \\
    \midrule
    \multicolumn{2}{@{}l@{}}{\emph{Frozen-append (Sec.~\ref{sec:supp-B-fa})}} \\
    Trains & \emph{New rows only}: the 80 stock rows and the stock no-object row are frozen, and only the $K$ new rows and their biases train. \emph{TFA from stock}: every row trains, and the stock and no-object rows start from the stock weights. \emph{LP, cosine head + cut}: exactly the baselines above, with their own $K{+}1$ softmax and the grids above. \\
    Data & The training images of the set; \emph{New rows, COCO neg.} adds the five COCO training images per class as extra data (399 images; every annotated object is a target). \\
    Offset & Every method sweeps a bias offset, one number subtracted from the biases of the $K$ new rows, over $\{0,1,2,3,4,6,8,12,16,24,32,48,64\}$. \\
    \bottomrule
  \end{tabular}
  \caption{\textbf{Grid of every method.}
  One fixed grid per method, the same for every shot count and seed; only the base-row choice of the \oursF{} depends on the set.}
  \label{tab:supp-grids}
\end{table}

\subsection{Batch orders of full-decoder fine-tuning}
\label{sec:supp-B-orders}

Full-decoder fine-tuning (FT) at five images per class is sensitive to the order in which the images reach the optimizer, so we run three batch orders per data seed, with the order independent of the seed.
Over the nine cells of seeds 0 to 2 and three orders, the standard deviation of its five-shot F1 is \adamfullCarsNfiveFoneCellsStd{} on cars, \adamfullCubNfiveFoneCellsStd{} on CUB, \adamfullPetsNfiveFoneCellsStd{} on Pets and \adamfullAircraftNfiveFoneCellsStd{} on aircraft.
The mean over the three orders is the number we compare with; the best order takes, for every seed, the order with the highest F1 on the test images, so it uses test labels that no other method uses.

\subsection{The 20-image COCO guard and COCO retention}
\label{sec:supp-B-guard}

The \oursF{}, the closed-form heads with the gate of Eq.~\ref{eq:formula} of the main paper and the frozen-append heads test the grid points in order of cross-validation score, best first, and keep the first candidate whose drop on the 20 images of the COCO guard (\Msec{sec:method-harness}) is within one point.
If no candidate passes, they keep the one with the smallest drop.
The image-only baselines (Prototype, LP, Tip-Adapter, the cosine head, FT and LwF) use no COCO~\cite{coco} images, and we report their COCO mAP drop on images that no method used for selection (images 21 to 200 and 201 to 400).
Full-decoder fine-tuning and LwF change the decoder, so we re-encode the COCO images with each fine-tuned decoder and score the stock rows plus the new rows on the result.

\subsection{Frozen-append}
\label{sec:supp-B-fa}

New rows only, new rows with COCO negatives and TFA from stock weights work in the $80{+}K{+}1$ softmax (rows 0 to 79 stock, the $K$ new rows, the no-object row last), and LP and the cosine head with a class-bias offset keep their own $K{+}1$ softmax; all five methods use the cached per-query embeddings of the other head baselines.
A candidate is a setting and a bias offset on the new rows; we rank the candidates by pooled 5-fold cross-validation macro-F1 on the training images and apply the guard as in Sec.~\ref{sec:supp-B-guard}.
The main ranking scores the cross-validation in the $K{+}1$ head, the same score the \oursF{} uses, and a second ranking scores it inside the $80{+}K{+}1$ softmax, which favors the baseline and matches what joint F1 measures.
The main paper reports, for every set, the best of the \faVariants{} guarded variants (five methods, two rankings) by mean five-shot F1, selected using the test labels.
In Tab.~\ref{tab:supp-fa}, new rows only, new rows with COCO negatives and TFA from stock weights meet the guard in \faPassedFa{}, \faPassedFaNeg{} and \faPassedTfaStock{} of \faPassedTotal{} runs, and LP and the cosine head with a class-bias offset mostly cannot meet it (\faPassedCtwoCut{} and \faPassedCfiveCut{} of \faPassedTotal{} runs).
Ranking in the 90-row head instead of the $K{+}1$ head (second block) raises the 90-row F1 of the new-rows methods and lowers their $K{+}1$ F1, and it leaves the result as it is: the \oursF{} reaches a higher F1 than the best head by this ranking on \faFormWinsJ{} of four sets and ties with it on the rest.

\begin{table}[t]
  \centering
  \small
\begingroup\setlength{\tabcolsep}{3pt}
\begin{adjustbox}{max width=\linewidth}
\begin{tabular}{lccccccccc}
\toprule
 & \multicolumn{5}{c}{F1 K+1} & F1 & COCO & parent & guard \\
\cmidrule(lr){2-6}
Method & Cars & CUB & Pets & Aircraft & mean & 90-row & lost & AP & met \\
\midrule
\multicolumn{10}{l}{\scriptsize\emph{Reference}} \\
Stock head & \na & \na & \na & \na & \na & \na & 0.0 & 65.2 & \na \\
\ours{} (Opus) & 58.1 & 32.4 & 52.3 & 23.6 & 41.6 & 30.5 & 0.8 & 43.8 & \na \\
\oursF{} & 65.0 & 39.6 & 61.7 & 22.3 & 47.2 & 42.0 & 0.9 & 52.2 & \na \\
\multicolumn{10}{l}{\scriptsize\emph{Guarded, selection in the K+1 head}} \\
New rows only & 36.2 & 14.6 & 21.4 & 20.8 & 23.2 & 9.7 & 0.9 & 62.5 & 20/20 \\
New rows, COCO negatives & 40.0 & 21.2 & 24.7 & 22.4 & 27.1 & 18.9 & 0.1 & 63.4 & 20/20 \\
TFA from stock weights & 0.0 & 0.0 & 0.3 & 0.3 & 0.1 & 0.1 & 1.3 & 61.3 & 20/20 \\
LP + class-bias cut & 0.3 & 0.0 & 0.0 & 3.0 & 0.8 & 0.0 & 2.8 & 63.4 & 0/20 \\
Cosine head + class-bias cut & 7.3 & 0.0 & 6.9 & 2.0 & 4.0 & 2.1 & 2.8 & 62.1 & 6/20 \\
\multicolumn{10}{l}{\scriptsize\emph{Guarded, selection in the 90-row head}} \\
New rows only & 33.0 & 15.3 & 20.3 & 17.1 & 21.4 & 16.2 & 0.8 & 59.9 & 20/20 \\
New rows, COCO negatives & 34.7 & 16.8 & 19.8 & 22.7 & 23.5 & 21.2 & 0.1 & 63.6 & 20/20 \\
TFA from stock weights & 0.0 & 0.0 & 0.6 & 0.3 & 0.2 & 0.1 & 1.3 & 60.5 & 20/20 \\
LP + class-bias cut & 0.3 & 0.0 & 0.0 & 3.0 & 0.8 & 0.0 & 2.8 & 63.4 & 0/20 \\
Cosine head + class-bias cut & 7.3 & 0.0 & 6.9 & 2.0 & 4.0 & 2.1 & 2.8 & 62.1 & 6/20 \\
\bottomrule
\end{tabular}
\end{adjustbox}
\par\vspace{2pt}{\scriptsize Five shots, seeds 0 to 4, all four fine-grained sets; F1 $\times$100. F1 K+1: the new rows alone; F1 90-row: the ten new classes inside the 80+10+1 softmax. COCO lost: mask-mAP points on val images 201 to 400. Parent AP: the task's COCO class after the edit, on val images 201 to 400 (stock 65.2; the stock row and the two reference rows are scored on the same images). Guard met: runs whose drop on the 20 guard images is within 1 point, of all runs (four sets, five seeds). The ranking scores each candidate (settings and class-bias cut) by 5-fold cross-validation on the train images; the first candidate that meets the guard is kept, else the smallest drop.}
\endgroup
  \caption{\textbf{Frozen-append heads at five shots.}
  The reference rows are the main runs of the main paper.
  A dash marks a cell that does not apply: the stock head has no new rows to score, and the guard count is given for the frozen-append methods only.}
  \label{tab:supp-fa}
\end{table}

\section{Harness and task prompt}
\label{sec:supp-C}

The harness is independent of the language model, so the same code serves every agent in the paper.

\subsection{Rules and checks}
\label{sec:supp-C-rules}

Everything outside the editable set (\Msec{sec:method-setup}) is frozen, and \texttt{save\_edit} rejects any edit to it.
The tool rejects unknown names, a wrong shape or dtype, values that are NaN or infinite, a tensor that equals the original times one constant, and edits that change nothing.
Before a code block runs, a static check parses it and rejects the block when it imports a denied module (file, process, network and introspection modules, and the packages of the model itself), calls a denied attribute (\texttt{backward}, \texttt{requires\_grad\_}, anything under \texttt{autograd}, \texttt{optim} or \texttt{torch.func}), loads or calls a model, or builds a name by string tricks to bypass the list.
While a block runs, an audit hook rejects writes outside the scratch folder, and while the program runs it also rejects reads outside it.
The rule against fitting weights to labels is checked by audit: we log every code block word for word and audit the log (Sec.~\ref{sec:supp-D}).
The checks stop accidental rule violations and obvious shortcuts.

\subsection{Program and task prompt}
\label{sec:supp-C-prompt}

At \texttt{final}, the harness runs the agent's program \texttt{recipe.py} on the run's own data and checks that it returns the saved edit to within $10^{-5}$, and every program in the paper passes this check.
The runs on the fine-grained sets, the ships, CD-FSOD and CLIP write a program; the PMT nuclei runs and the SAM~3 runs of Sec.~\ref{sec:supp-H} do not.
The listing below is the prompt of one five-shot cars run, with file paths replaced by \texttt{\$RUN} and the class names and the three sentences that describe the images and the classes shortened; a line in square brackets stands for a part we left out and says what it holds.
It gives the parts on the task, the scoring, the rules and the working steps; the full prompt, with the tool reference (the kernel, the forward tool of \Msec{sec:method-harness}, a status call and the \texttt{final} call), the description of the head and the instructions for the program, is part of the code release.
Other sets change only the project line, the class names, the sentences on the images and the classes, the scoring sentence and the head description.

\begin{lstlisting}[language={},basicstyle=\ttfamily\fontsize{5.5pt}{6.2pt}\selectfont,breaklines=true,breakindent=4pt,columns=fullflexible]
# TASK

You are given the class head of PMT, an instance segmentation model (a frozen DINOv3 image encoder and a mask decoder) that returns a mask, a class and a score for each object it finds. The project is "stanford-cars-10". Its classes: <the ten class names>.
<three sentences on the images and the classes>. Scored with box IoU (the box around a predicted mask).

Your goal: make this head better at the project's task, as measured by instance macro-F1 on held-out images.

You do it by EDITING THE WEIGHTS DIRECTLY. Find out what goes wrong, work out which parts of the head are responsible, and change them in a way you can justify with evidence. No training.

# HOW IT IS MEASURED

After you finish, your edited head and the original head are scored on held-out images of this project. You never see those images or their labels. The score is instance macro-F1: per class, predicted instances are matched to true instances at IoU 0.5 and F1 is computed; then it is averaged over classes. Only weight values can change: same tensor names, shapes and float32 dtype.

# WHAT YOU HAVE

[the tool reference (exec, fwd generate / read / retain, status)]

# THE HEAD

[the frozen parts, the editable tensors, the scoring rule and the read request]

# RULES

Not allowed:
(a) Gradient descent or any other training: no backward passes, autograd, optimizers, requires_grad, torch.func gradients.
(b) Iterative fitting of weights to a loss or to labels, and closed-form regression or least-squares fits of weights to labels. You may analyse, measure and compute simple statistics, and make edits you can explain (zero or shift a unit, remove a direction, change a bias, swap rows, set one value from a measured number). Edits must come from understanding, not from fitting a model to the labels.
(c) Running the head yourself in the kernel (head_forward, loading a model, re-implementing its forward pass on images). Only the fwd tool runs it.
(d) Importing os, subprocess, socket, urllib, requests and similar, opening files outside WORK, internet access.
(e) Whole-tensor rescales and edits to frozen tensors (refused by save_edit).
(f) Breaking what the model already does: keep retention within 1 point of the original (the retain check). Run it on your final edits before you call final.
Code that breaks (a), (c) or (d) is refused before it runs. These checks are not perfect: every code block you run is logged word for word and will be audited, so follow the rules in spirit as well.

# HOW TO WORK

1. Diagnose first: run `generate` on the original head, look at per-class and per-image F1, find the biggest failure type you can explain (missed instances, false positives, wrong class, duplicates).
2. Use `read` to see what the head does on those cases; show that a part causes the failure before you edit it. Try several edit strengths.
3. Compare original and edited on the SAME images ("model": "both"). Keep an edit only if the gain is clearly larger than its standard error. Confirm on images you did not use to choose the edit when the image pool is big enough; when it is small, only keep edits you can explain, and check that most images improve (the paired result counts better and worse images).
4. Plan your 25 passes. State the hypothesis of each code run in its first comment line, print compact summaries, never whole arrays.
5. Keep notes in `$RUN/work/notes.md`.

# HOW IT ENDS

[the final call and the program recipe.py]
\end{lstlisting}

\subsection{Data, runs and code}
\label{sec:supp-C-repro}

The manifests of the detection sets built for this work, with the training lists, are part of the code release.
Every run of the agent keeps its prompt, a log of every code block and tool call word for word, and its edited tensors; a run that writes a program also keeps its \texttt{recipe.py} and a record of the checks at \texttt{final}.
The code release holds the harness, the backend for the PMT head, the scorer, the \oursF{} and its COCO guard, the replay of a program, the paired bootstrap, and three of the gradient baselines (LP, the cosine head, and FT with LwF).
A rerun needs the reader's own agent API key and the model weights.

\section{Fitting audit}
\label{sec:supp-D}

\subsection{What the audit counts}
\label{sec:supp-D-audit}

The audit scans, with regular expressions, the log of each of the \auditRunsTotal{} Opus runs on seeds 0 to 2 at 5, 10 and 20 shots, with both starting rows on CUB and Pets (Sec.~\ref{sec:supp-E-main}).
\auditRunsFit{} of \auditRunsTotal{} runs call a least-squares, pseudo-inverse, SVD or eigen-decomposition routine at least once, mostly to compute the statistics that the prompt allows, namely class means and the principal directions that the program removes.
A regular-expression scan treats such a statistic and a fit of weights to labels alike, so we report the programs with an LDA-style metric apart (Sec.~\ref{sec:supp-D-nolda}).
\auditRunsHand{} runs write a literal hand-set number into the edit.
The agents choose between edits by training F1, which is the label use that \Msec{sec:method-harness} states, and every baseline uses the same labels.

\subsection{Without the LDA-style programs}
\label{sec:supp-D-nolda}

Tab.~\ref{tab:supp-nolda} repeats the main comparison of \Msec{sec:res-main} without the runs whose program uses a pooled inverse covariance, the LDA-style metric that the rule against fitting does not clearly separate from a fit (\Msec{sec:method-harness}): of the \noldaFlagged{} flagged programs, \noldaHeadlineFive{} are main five-shot runs, so \noldaFiveUnchanged{} of \noldaFiveCells{} five-shot comparisons stay unchanged.
At ten shots \noldaHeadlineTen{} main runs are excluded, and the gap to FT's best order moves by at most \noldaTenMove{} points.

\begin{table}[t]
  \centering
  \small
\begingroup\setlength{\tabcolsep}{3pt}
\begin{adjustbox}{max width=\linewidth}
\begin{tabular}{lccccccc}
\toprule
 & & \multicolumn{2}{c}{Agent edit} & \multicolumn{2}{c}{FT, best order} & \multicolumn{2}{c}{FT, mean of orders} \\
\cmidrule(lr){3-4}\cmidrule(lr){5-6}\cmidrule(lr){7-8}
Set & seeds out & all & kept & all & kept & all & kept \\
\midrule
\multicolumn{8}{l}{\scriptsize\emph{5 shots}} \\
Cars & 0 & 58.1 & 58.1 & 71.2 & 71.2 & 58.1 & 58.1 \\
CUB & 0 & 32.4 & 32.4 & 28.4 & 28.4 & 22.5 & 22.5 \\
Pets & 0 & 52.3 & 52.3 & 48.6 & 48.6 & 39.4 & 39.4 \\
Aircraft & 0 & 23.6 & 23.6 & 23.9 & 23.9 & 20.9 & 20.9 \\
\multicolumn{8}{l}{\scriptsize\emph{10 shots}} \\
Cars & 2 & 81.7 & 79.2 & 96.4 & 97.2 & 80.8 & 92.4 \\
CUB & 1 & 46.5 & 49.6 & 50.3 & 50.3 & 46.1 & 46.2 \\
Pets & 0 & 64.9 & 64.9 & 76.1 & 76.1 & 65.6 & 65.6 \\
Aircraft & 0 & 25.8 & 25.8 & 38.1 & 38.1 & 32.8 & 32.8 \\
\bottomrule
\end{tabular}
\end{adjustbox}
\par\vspace{2pt}{\scriptsize F1 $\times$100, plain mean over seeds. ``kept'': the seeds of the main runs whose program uses a pooled inverse-covariance (LDA-style) metric are dropped from all three columns of that cell (the FT numbers move because they are averaged over the remaining seeds only). Seeds out: how many that is. 30 Opus recipes are flagged; none of them is a main five-shot run, so the five-shot rows do not change.}
\endgroup
  \caption{\textbf{Main comparison without the LDA-style programs.}
  \emph{All} uses every seed; \emph{kept} drops, from all three methods, the seeds whose \ours{} program uses a pooled inverse covariance.}
  \label{tab:supp-nolda}
\end{table}

\section{Ablations and controls}
\label{sec:supp-E}

\subsection{Pictures, budget and starting row}
\label{sec:supp-E-main}

Tab.~\ref{tab:supp-abl5} gives the five-shot results, where each row changes one thing against the first row, the setting of the main paper: pictures of the errors on, 25 forward passes, Claude Opus~5.5.
With pictures on, every \emph{generate} request also draws the errors on the images it scored, with outlines for correct masks, missed true instances, false masks and masks of the wrong class.
The runs without pictures score as high as the runs with them on all four sets, and the gaps lie inside the spread over seeds.
Ten passes and 50 passes give means within the seed spread of 25 passes.
For CUB and Pets every variant in the table uses the parent start, in which the first new row holds the COCO bird or cat row (\Msec{sec:method-setup}), and that row already separates the object from the background.
With the first ten stock rows as the start, whose meaning is unrelated to the task, the \ours{} scores \opusOvlPetsNfive{} against \meanthree{opusOvlParPetsNfive} F1 on Pets and \opusOvlCubNfive{} against \meanthree{opusOvlParCubNfive} on CUB (three seeds).

\begin{table}[t]
  \centering
  \small
\begingroup\setlength{\tabcolsep}{3pt}
\begin{adjustbox}{max width=\linewidth}
\begin{tabular}{lcccc}
\toprule
Variant & Cars & CUB & Pets & Aircraft \\
\midrule
Opus, pictures on, $B{=}25$ & 55.4\pmstd{14.8} & 32.0\pmstd{6.5} & 55.1\pmstd{8.5} & 22.4\pmstd{5.2} \\
\quad pictures off & 59.6\pmstd{3.8} & 33.1\pmstd{6.6} & 57.5\pmstd{3.5} & 23.5\pmstd{4.7} \\
\quad $B{=}10$ & 50.4\pmstd{14.2} & 32.0\pmstd{8.5} & 53.5\pmstd{5.8} & 24.1\pmstd{3.0} \\
\quad $B{=}50$ & 54.5\pmstd{17.2} & 29.9\pmstd{2.1} & 54.1\pmstd{5.8} & 26.4\pmstd{2.6} \\
\midrule
Sol & 41.0\pmstd{4.5} & 34.0\pmstd{4.3} & 52.4\pmstd{3.9} & 22.2\pmstd{2.5} \\
Sonnet & 5.1\pmstd{2.4} & 2.7\pmstd{0.6} & 4.2\pmstd{2.1} & 3.7\pmstd{2.8} \\
\bottomrule
\end{tabular}
\end{adjustbox}
\par\vspace{2pt}{\scriptsize F1 $\times$100, five shots, mean$\pm$std over seeds 0 to 2.}
\endgroup
  \caption{\textbf{Ablations of the \ours{} at five shots.}
  Each row changes one thing against the first row; Sol and Sonnet use the same harness, budget and prompt as Opus.
  The first row is the main setting on seeds 0 to 2, so its means differ from the five-seed means of \Mtab{tab:main}.}
  \label{tab:supp-abl5}
\end{table}

\subsection{Agent model}
\label{sec:supp-E-agent}

Sonnet~5 fails in every cell of Tab.~\ref{tab:supp-abl5} (\Msec{sec:limits}).
The failure is not a harness fault: every Sonnet run ends with its own \texttt{final} call, most of them before the pass budget is spent, and each program reproduces its edit.
Its training F1 stays at \sonnetTrainLo--\sonnetTrainHi{}, so it already fails on the labeled images.
Most of its programs build class means, but none removes the directions that vary within a class, and several leave a hand-picked set of class rows unedited.

\subsection{Vision-language model on true crops}
\label{sec:supp-E-vlm}

We hand Claude Opus~5.5 the true crop of each test object and ask for its class, in three variants on the same query crops: \emph{in-context} (the ten class names and five support crops per class), \emph{zero-shot} (names, no support crops), and \emph{hidden names} (the same support crops, relabeled as groups with a random mapping per seed, so no real name is visible anywhere).
On the four fine-grained sets of Tab.~\ref{tab:supp-vlm}, zero-shot accuracy is within a few points of the in-context arm, so most of the score comes from what the model learned in pretraining about these datasets.
With the names hidden the accuracy stays close to or above the in-context variant on these four sets, so the model can also solve the task by visual matching alone.
On LIVECell, where the model has little prior, zero-shot accuracy falls close to chance, and there the model learns from the five crops.
The heads we compare classify \headNameLo--\headNameHi\% of the objects they find correctly, against \vlmCropLo--\vlmCropHi\% for the vision-language model on the four fine-grained sets, so the gap is in classifying the object and not in finding it, and a model that is given the true box is not a detector (\Msec{sec:limits}).
The vision-language model costs \$\vlmCostLo--\$\vlmCostHi{} and \vlmSecLo--\vlmSecHi{} seconds per query, and the edited head needs \headMsPerImage{} ms per image on CPU and no network call.

\begin{table}[t]
  \centering
  \small
  \begin{tabular}{lccc}
    \toprule
    Set & In-context & Zero-shot & Hidden names \\
    \midrule
    Cars     & \mval{vlmiclCarsNfiveCropAcc}     & \mval{vlmiclZeroshotCarsNfiveCropAcc}     & \mval{vlmiclHidCarsNfiveCropAcc} \\
    CUB      & \mval{vlmiclCubNfiveCropAcc}      & \mval{vlmiclZeroshotCubNfiveCropAcc}      & \mval{vlmiclHidCubNfiveCropAcc} \\
    Pets     & \mval{vlmiclPetsNfiveCropAcc}     & \mval{vlmiclZeroshotPetsNfiveCropAcc}     & \mval{vlmiclHidPetsNfiveCropAcc} \\
    Aircraft & \mval{vlmiclAircraftNfiveCropAcc} & \mval{vlmiclZeroshotAircraftNfiveCropAcc} & \mval{vlmiclHidAircraftNfiveCropAcc} \\
    Ships    & \mval{vlmiclXviewNfiveCropAcc}    & \mval{vlmiclZeroshotXviewNfiveCropAcc}    & \mval{vlmiclHidXviewNfiveCropAcc} \\
    LIVECell & \mval{vlmiclPmtLivecellNfiveCropAcc} & \mval{vlmiclZeroshotPmtLivecellNfiveCropAcc} & \mval{vlmiclHidPmtLivecellNfiveCropAcc} \\
    \bottomrule
  \end{tabular}
  \caption{\textbf{Vision-language model handed the true crop.}
  Top-1 class accuracy (\%) at five support crops per class, mean over three seeds.
  This is class accuracy on given objects and differs from detection F1.
  LIVECell is a set of phase-contrast cell lines on which the model has little prior.}
  \label{tab:supp-vlm}
\end{table}

\subsection{Cost}
\label{sec:supp-E-cost}

An agent run on PMT takes \agentMinLo--\agentMinHi{} minutes of wall time and \$\agentUsdLo--\$\agentUsdHi{} of model API cost per task.
The \oursF{} needs \formCpuLo--\formCpuHi{} CPU seconds per cell and no GPU.
FT with its grid needs \cfGpuLo--\cfGpuHi{} GPU minutes per cell, which counts all five folds and the refit.
The agent ranges are over the cells of set and shot count at five to 20 shots; the CPU and GPU ranges go from the mean over the sets at five shots to the mean at 20 shots.

\section{Programs}
\label{sec:supp-F}

\subsection{Two programs}
\label{sec:supp-F-listings}

The listings come from two five-shot runs of the \ours{}, one for each kind of start: cars, from the COCO car row, and CUB, from the bird row copied into the first new row.
We cut set-up lines and comments (marked \texttt{...}), write a few library calls in short form (\texttt{svd}, \texttt{norm}, \texttt{class\_means}), and keep the lines that build the rows; \texttt{E} holds the query embeddings, \texttt{prim} marks the primary query of each labeled object, and \texttt{Wn} and \texttt{bn} are the new rows and their biases.
Each one builds rows from class means of query embeddings, removes directions that vary between objects of one kind, and sets a gate or a bias so that a row gives a confident detection on the one query that holds the object.

\noindent\textbf{Cars.}
The program writes each class row as the COCO car row plus a gate direction plus a class direction.
\texttt{SPEC} is its read request, \texttt{mc} the true class of each query (positive only on a labeled car), \texttt{carp} the original car score of each query, \texttt{K} the number of removed directions, \texttt{half} half the gap between the lowest gate score of a labeled car and the highest of a background car, and \texttt{boff} a bias offset set so that the weakest labeled training car gets an added logit of \texttt{DELTA0} on its top class.
\begin{lstlisting}
K, ALPHA, GAMMA, DELTA0 = 10, 20.0, 20.0, 5.0
def edit(train, weights):
    r = train.read(SPEC)
    ...
    prim = ...  # per image: matched query with the
                # highest original car score
    bgc = (mc <= 0) & (carp >= 0.5)  # unlabelled cars
    # gate: labelled cars vs background cars
    u = E[prim].mean(0) - E[bgc].mean(0)
    t = 0.5*((E[prim]@u).min() + (E[bgc]@u).max())
    # class directions, top-K car PCs removed
    mu0 = E[prim | bgc].mean(0)
    Vk = svd(E[prim | bgc] - mu0).Vt[:K]
    X = E[prim] - mu0;  X -= (X @ Vk.T) @ Vk
    C = class_means(X / norm(X));  C -= C.mean(0)
    ...
    for c in range(NCLS):
        Wn[c] = W0[CAR_ROW] + (GAMMA/half)*u + ALPHA*C[c]
        bn[c] = b0[CAR_ROW] - GAMMA*t/half \
                - ALPHA*(C[c] @ mu0) + boff
\end{lstlisting}

\noindent\textbf{CUB, bird-row start.}
Each class row is the bird row plus a species direction.
The program removes within-class and all-query directions, and a percentile \texttt{K} of the class term on non-object queries sets the bias.
\begin{lstlisting}
ALPHA, M, KW, KB = 5.0, -4.0, 5, 80
def edit(train, weights):
    ...
    # primary query: best query on a true bird
    X = E[prim];  mus = class_means(X, y)
    m = mus[present].mean(0)
    # nuisance: within-class PCs + top PCs of all queries
    Q = qr(cat([svd(X - mus[y]).Vt[:KW],
                svd(E - E.mean(0)).Vt[:KB]]))
    B[c] = (mus[c] - m) - proj(mus[c] - m, Q)
    # class term must stay below object queries
    T = (E - m) @ B.T - 0.5*n2c
    K = percentile(T[non_primary].max(1), 99)
    Wn[c] = wb + ALPHA*B[c]
    bn[c] = bb - ALPHA*(B[c] @ m + 0.5*n2c[c]) \
            - ALPHA*K + M
\end{lstlisting}

\subsection{The formula as code}
\label{sec:supp-F-formula}

Fig.~\ref{fig:supp-formula} shows the core of the formula (Eq.~\ref{eq:formula} of the main paper).
\texttt{t\_gate}, \texttt{s\_cls} and \texttt{k\_pcs} are $g$, $s$ and $m$, and \texttt{cut} is the offset $c$.

\begin{figure}[t]
  \centering
\noindent\fcolorbox{black!45}{white}{\begin{minipage}{\dimexpr\linewidth-2\fboxsep-2\fboxrule\relax}
\ttfamily\footnotesize\frenchspacing\raggedright\setlength{\parindent}{0pt}\setlength{\baselineskip}{9.4pt}%
\textcolor{black!60}{\# Xp: one primary query per object, Xo: rest}\\
d\ \ = Xp.mean(0) - Xo.mean(0)\\
v\ \ = d / d.norm() \textcolor{black!60}{\# or d / (Xp@d).std()}\\
th\ = ((Xp@v).mean() + (Xo@v).mean()) / 2\\
mu\ = class means of Xp;\ m0 = Xp.mean(0)\\
Dc\ = mu - mu.mean(0)\\
Dc -= (Dc @ Vh.T) @ Vh \textcolor{black!60}{\# k\_pcs nuisance PCs}\\
bD\ = -(Dc @ m0) - 0.5*(Dc**2).sum(1)\\
W[c] = Wb + t\_gate*v + s\_cls*Dc[c]\\
b[c] = bb - t\_gate*th + s\_cls*bD[c]\\
b[:K] -= cut\\
\textcolor{black!60}{\# Wb, bb: parent row, or no-object row}
\end{minipage}}
  \caption{\textbf{Core of the formula.}
  \texttt{Xp} holds the primary queries of the labeled objects and \texttt{Xo} the other queries; \texttt{Wb} and \texttt{bb} are the base row; \texttt{Vh} holds the top \texttt{k\_pcs} directions of the within-class residuals of \texttt{Xp}, and \texttt{K} is the number of classes.}
  \label{fig:supp-formula}
\end{figure}

\subsection{How the terms were counted}
\label{sec:supp-F-converge}

Panel (c) of \Mtab{tab:program} counts how often the agents' programs contain the three terms of the formula (\Msec{sec:method-out}).
The counts cover the independent programs, those of runs that were not handed another run's program: every such finished Opus program on the four fine-grained sets, their invented-class variants and the ship set.
A rule-based scan of the code assigns each program a flag for each of the terms below, and we checked a sample of the flags by hand, so the counts are estimates.
\begin{itemize}\itemsep1pt
\item \emph{Class-mean rows:} the rows are built from per-class means of query embeddings.
\item \emph{Removal of shared directions:} an SVD, eigen-decomposition or QR projection drops the top directions, or an inverse-covariance (whitened, LDA-style) metric is used.
\item \emph{Explicit gate:} a learned primary-versus-other direction is added to the class rows, or the stock parent-versus-no-object margin is used to increase the margin of the parent row over the no-object row.
\item \emph{Exact formula shape:} a learned gate direction, class means and removal of within-class directions.
\end{itemize}
First, every prompt has the same instructions for the program (in the full prompt; Sec.~\ref{sec:supp-C-prompt}), and they name ``class means, directions, thresholds'', so part of the overlap comes from their vocabulary and from what the tools expose; we therefore say that the terms recur in the programs.
Second, class means are close to universal because they are the main statistic the rules allow, so the informative terms are the gate and the removal of directions.
Third, the gate is the weakest term: in parent-start runs on CUB and Pets, the parent row and a bias shift do its work, and few of these programs learn a gate direction.

\subsection{A second draw of classes}
\label{sec:supp-F-fresh}

To test whether the formula was shaped on the classes we looked at, we drew ten new classes for each of the four sets, none from the first draw: ten other sedans, ten sparrows, ten dog breeds, and ten other Boeing and Airbus variants.
The \oursF{} and its grid were frozen before the draw, and every method ran the same five-shot cells on five seeds (Tab.~\ref{tab:supp-fresh10}).
On the ten sparrows it scores lower F1 than FT's best order (difference \freshBestDiffCub{} F1 points, 95\% interval \freshBestLoCub{} to \freshBestHiCub{}), and its COCO guard fails in \freshGuardFailed{} of the \freshGuardCells{} cells of this draw, where it keeps the candidate with the smallest COCO drop.

\begin{table}[t]
  \centering
  \small
\begingroup\setlength{\tabcolsep}{3pt}
\begin{adjustbox}{max width=\linewidth}
\begin{tabular}{lcccccc}
\toprule
 & \multicolumn{4}{c}{F1 $\times$100} & \multicolumn{2}{c}{\oursF{} minus FT} \\
\cmidrule(lr){2-5}\cmidrule(lr){6-7}
Set & \oursF{} & FT mean & FT best & LP & mean order & best order \\
\midrule
Cars & 55.7\pmstd{4.2} & 44.7\pmstd{16.0} & 58.7\pmstd{17.2} & 26.5\pmstd{3.3} & +8.6 [\ensuremath{-}3.6, +19.9] & \ensuremath{-}3.2 [\ensuremath{-}16.3, +9.5] \\
CUB & 24.0\pmstd{3.0} & 29.9\pmstd{13.4} & 38.8\pmstd{13.4} & 14.8\pmstd{3.4} & \ensuremath{-}7.5 [\ensuremath{-}17.5, +4.3] & \ensuremath{-}12.6 [\ensuremath{-}26.0, \ensuremath{-}0.1]$\downarrow$ \\
Pets & 74.3\pmstd{4.7} & 60.8\pmstd{11.7} & 72.8\pmstd{7.2} & 20.8\pmstd{2.6} & \textbf{+13.3 [+3.1, +23.4]} & +1.2 [\ensuremath{-}6.4, +9.7] \\
Aircraft & 50.7\pmstd{4.0} & 45.5\pmstd{4.8} & 52.0\pmstd{4.7} & 41.1\pmstd{4.8} & +4.3 [\ensuremath{-}2.5, +11.8] & \ensuremath{-}1.5 [\ensuremath{-}9.1, +6.5] \\
\bottomrule
\end{tabular}
\end{adjustbox}
\par\vspace{2pt}{\scriptsize Ten new classes per set, none from the first draw; formula and grid as frozen before the draw. Mean$\pm$std over five seeds. \oursF{} minus FT: F1 points, where each method's macro-F1 is computed on its counts summed over the five seeds (FT mean order: the mean of that F1 over the three orders), so the seed-mean columns do not subtract to it; two-level 95\% interval; bold: interval above 0; $\downarrow$: below 0. The formula's 20-image COCO guard failed in 4 of 20 cells.}
\endgroup
  \caption{\textbf{The \oursF{} on a second draw of ten classes per set.}
  Five shots, five seeds, with FT at its mean and its best batch order and LP.}
  \label{tab:supp-fresh10}
\end{table}

\section{Qualitative examples}
\label{sec:supp-G}

\begin{figure*}[t]
  \centering
  \includegraphics[width=\textwidth]{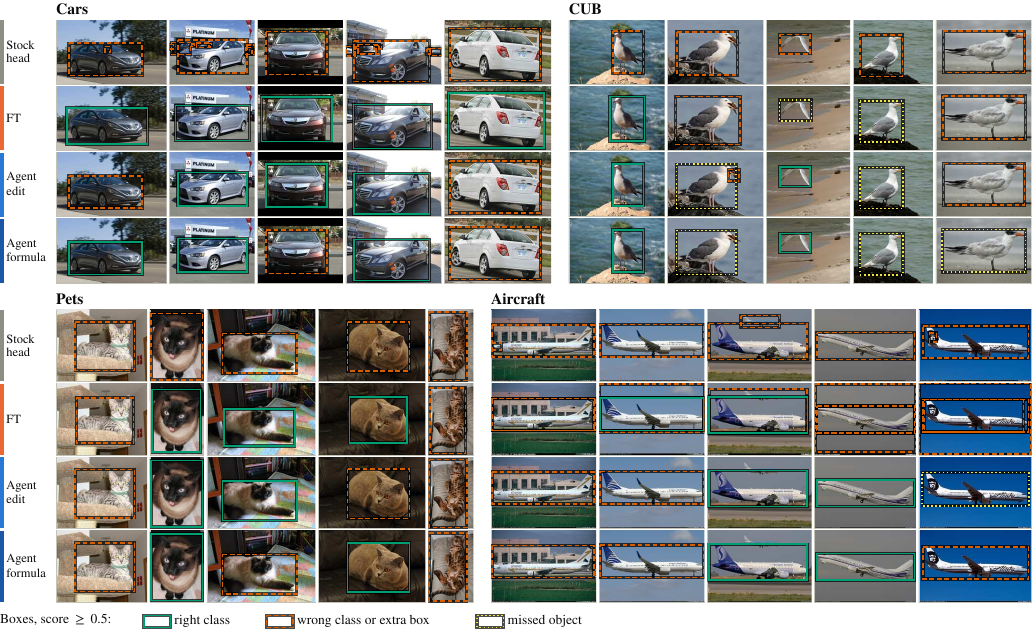}
  \caption{\textbf{Detections on the same test photos.}
  Five shots, seed~0, score threshold 0.5; each photo holds one labeled object.
  Rows: the stock head that the agent starts from, FT at its best batch order for this seed, the \ours{} and the \oursF{}.
  Solid green: a box of the right class that matches an unused true instance at box IoU~0.5, on every set (the tables score pets with mask IoU).
  Dashed orange: a box of the wrong class, or an extra box.
  Dotted yellow: a true object that no box covers.
  Photos: the first five of a random draw per set, fixed before any detection was viewed.}
  \label{fig:supp-qual}
\end{figure*}

Fig.~\ref{fig:supp-qual} shows the same test photos under the stock head, full-decoder fine-tuning (FT), the \ours{} and the \oursF{} at five shots.
We drew the test photos of each set at random before we looked at any detection, and we show the first five of the draw in its order, with none replaced.
FT is shown at the batch order with the highest test F1 for this seed, which favors it.
In these photos FT is right most often on cars, and the three learned heads are close on the other sets.

\section{SAM~3 and the gate}
\label{sec:supp-H}

\subsection{Gate part and class part}
\label{sec:supp-H-gain}

\begin{figure}[t]
  \centering
  \includegraphics[width=\linewidth]{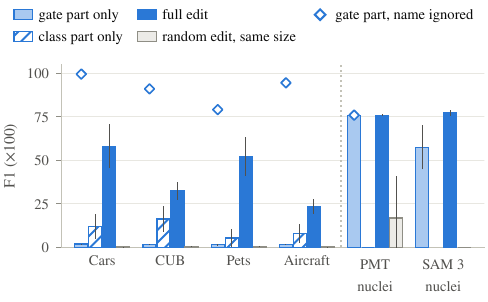}
  \caption{\textbf{Decomposition of the gain.}
  F1 ($\times$100) of the gate part alone, the class part alone, the full edit and a random edit of the same size, for PMT on the four fine-grained sets (\ours{}, five shots) and for PMT and SAM~3 on nuclei (one class, 20 shots; GPT~6.1 Sol on PMT, Opus and Sol on SAM~3).
  Diamonds score the gate part as finding the object without assigning its class (class-merged F1); bars: std over runs.}
  \label{fig:supp-gain}
\end{figure}

An edit can act as a gate, which decides whether an object is present, or add knowledge of the classes.
Most programs write both into the class rows, so we split inside the rows (Fig.~\ref{fig:supp-gain}): the shared part, which acts as the gate, is the mean change of the new rows plus the whole change of the no-object row, and the class part is each row minus the shared part.
On the fine-grained sets the shared part finds the object (the diamonds, which score it without its class) but cannot assign its class, with macro-F1 of \gateMacroLo--\gateMacroHi, and the class part alone scores far below the full edit, so both parts are needed.
On the fine-grained sets, a random edit of the same size scores below \randomEditMax{} F1.
On PMT nuclei the gate part equals the full edit, and on SAM~3 nuclei the gate part recovers most of it.
With one class there is nothing class-specific to learn.

\subsection{SAM~3 setup}
\label{sec:supp-H-setup}

SAM~3~\cite{sam3} is a text-prompted detector, and the agent edits the decoder and the heads after text-image fusion in the harness of Sec.~\ref{sec:supp-C}.
We use four sets with 20 training images and \nEvalHard{} evaluation images each (\nEvalIsaid{} on iSAID): three hard domains, cell lines in phase-contrast images (LIVECell~\cite{livecell}), nuclei with one foreground class (BBBC038~\cite{bbbc038}) and aerial images (iSAID~\cite{isaid}), and five COCO vehicle classes as a control where the stock model already works.
The reference method is \emph{a per-project trained head}: a head fit with gradients on the same 20 images for each project.
The agents are Claude Opus~5.5 and GPT~6.1 Sol, with two seeds each.

\subsection{SAM~3 results}
\label{sec:supp-H-results}

\begin{table}[t]
  \centering
  \small
\begingroup
\begin{adjustbox}{max width=\linewidth}
\begin{tabular}{lccccc}
\toprule
Set & Stock & Trained head & Adam & \ours{} (Opus) & \ours{} (Sol) \\
\midrule
LIVECell & 0.0 & \textbf{25.8}\pmstd{0.3} & 0.0 & 20.7\pmstd{2.9} & 14.1\pmstd{0.5} \\
BBBC038 & 0.0 & \textbf{84.9}\pmstd{0.7} & 71.8 & 76.7\pmstd{1.6} & 78.0\pmstd{1.0} \\
iSAID & 9.2 & 18.4\pmstd{1.3} & 16.9 & \textbf{21.1}\pmstd{1.3} & 20.4\pmstd{0.3} \\
COCO-veh. & 76.6 & 48.4\pmstd{3.0} & 57.9 & 76.7\pmstd{0.8} & \textbf{77.5}\pmstd{0.9} \\
\bottomrule
\end{tabular}
\end{adjustbox}
\par\vspace{2pt}{\scriptsize SAM~3 macro-F1 ($\times$100), $N{=}20$ train images. Trained head: a per-project head fit with gradients on the same 20 images. Adam: one run (seed 0) on the same tensors the agent edits. Other columns: mean$\pm$std over seeds. Bold: best of the four right-hand columns.}
\endgroup
  \caption{\textbf{SAM~3 on three hard domains and COCO vehicles.}
  The trained head has three seeds and each agent two; the stock model does not depend on the training draw.}
  \label{tab:supp-sam3}
\end{table}

On SAM~3, the stock model scores zero on LIVECell and BBBC038 (Tab.~\ref{tab:supp-sam3}): the stock presence logit puts every score below the score threshold, so no instance passes the threshold.
The \ours{} raises F1 on BBBC038 to \samNucleiLo--\samNucleiHi{} F1 across Opus and Sol, where a random edit of the same size and norm stays at \opusSamBbbcNtwentyRandom{} and Adam~\cite{adam} on the same tensors reaches \samNucleiAdam{}.
Adam on the same tensors is one run per set, which we read as a reference; on LIVECell it never raises the presence logit above the score threshold, so it scores \adamSamLivecellNtwenty.
A per-project trained head scores higher F1 than the edited SAM~3 on LIVECell and BBBC038, and the edit scores higher F1 than the trained head on iSAID and on COCO vehicles, where the trained head scores well below the stock model; these are point estimates.

\section{CLIP}
\label{sec:supp-I}

The harness drives a third head, the classifier of CLIP~\cite{clip}.

\noindent\textbf{Setup.}
We use CLIP ViT-B/16 on six classification sets: Oxford Pets~\cite{pets}, Stanford Cars~\cite{stanfordcars}, FGVC Aircraft~\cite{aircraft}, Flowers~\cite{flowers}, DTD~\cite{dtd} and EuroSAT~\cite{eurosat}, with the CoOp few-shot splits and 1, 4 and 16 shots per class.
The editable tensors are the classifier weights, the biases and the logit scale, and the starting weights are the zero-shot text rows.
Claude Opus~5.5 runs with a budget of 25 passes, pictures and a program (\texttt{recipe.py}) on three seeds, and the harness is the one of Sec.~\ref{sec:supp-C}.
The baselines are the linear probe, LP++~\cite{huang2024lpp}, Tip-Adapter and Tip-Adapter-F~\cite{zhang2022tipadapter} and GDA~\cite{wang2024gda}; each one tunes its settings on a few-shot validation list of at most four images per class, so it uses the same labels as the agent.
The published numbers of Tip-Adapter-F and GDA tune on larger validation splits and reach higher accuracy on some sets; we use the matched-label protocol.
The linear probe does not use the text rows, so at one shot it scores below zero-shot CLIP on Pets, Cars, Aircraft and DTD.

\begin{table*}[!t]
  \centering
  \footnotesize
\begingroup\setlength{\tabcolsep}{2.5pt}
\begin{adjustbox}{max width=\linewidth}
\begin{tabular}{lcccccccccccccccccc}
\toprule
 & \multicolumn{3}{c}{Pets} & \multicolumn{3}{c}{Cars} & \multicolumn{3}{c}{Aircraft} & \multicolumn{3}{c}{Flowers} & \multicolumn{3}{c}{DTD} & \multicolumn{3}{c}{EuroSAT} \\
\cmidrule(lr){2-4}\cmidrule(lr){5-7}\cmidrule(lr){8-10}\cmidrule(lr){11-13}\cmidrule(lr){14-16}\cmidrule(lr){17-19}
Shots & 1 & 4 & 16 & 1 & 4 & 16 & 1 & 4 & 16 & 1 & 4 & 16 & 1 & 4 & 16 & 1 & 4 & 16 \\
\midrule
Zero-shot CLIP & \multicolumn{3}{c}{89.1} & \multicolumn{3}{c}{65.5} & \multicolumn{3}{c}{24.8} & \multicolumn{3}{c}{70.8} & \multicolumn{3}{c}{44.1} & \multicolumn{3}{c}{48.3} \\
\midrule
Linear probe & 42.2 & 70.3 & 85.8 & 34.9 & 63.3 & 80.9 & 20.4 & 32.7 & 46.2 & 71.8 & 92.7 & 97.5 & 36.2 & 56.2 & 69.5 & 52.5 & 76.5 & 85.3 \\
Tip-Adapter & 89.3 & 90.0 & 91.6 & 66.8 & 70.3 & 74.8 & 27.9 & 32.7 & 39.1 & 83.3 & 91.1 & 93.7 & 49.8 & 58.1 & 65.2 & \textbf{63.0} & 70.4 & 72.9 \\
Tip-Adapter-F & 89.4 & 91.2 & 92.2 & \textbf{67.9} & 72.4 & 81.8 & 28.4 & 33.4 & 45.3 & 82.2 & 91.7 & 97.1 & 50.3 & 60.8 & 67.9 & 62.1 & 71.5 & 80.0 \\
GDA & 89.5 & 91.4 & \textbf{92.9} & 66.0 & \textbf{75.0} & 82.8 & 25.7 & \textbf{37.0} & \textbf{50.4} & 79.7 & \textbf{94.6} & \textbf{97.7} & 47.1 & 60.7 & \textbf{71.4} & 61.4 & \textbf{79.9} & \textbf{88.8} \\
LP++ & 89.8 & 91.0 & 92.7 & 67.2 & 74.2 & 80.2 & \textbf{28.9} & 34.0 & 42.0 & \textbf{85.0} & 93.4 & 96.1 & \textbf{51.3} & \textbf{62.0} & 71.1 & 62.9 & 74.9 & 85.3 \\
CoOp (published) & \textbf{90.2} & \textbf{92.5} & 92.0 & 67.5 & 74.4 & \textbf{82.9} & 20.8 & 30.9 & 43.2 & 78.3 & 92.2 & 96.8 & 50.1 & 59.5 & 69.7 & 56.4 & 69.7 & 85.0 \\
\ours{} (Opus) & 89.8 & 90.8 & 91.4 & 66.7 & 71.9 & 76.6 & 24.7 & 32.3 & 43.4 & 81.8 & 89.2 & 93.3 & 50.5 & 55.7 & 63.9 & 60.8 & 65.0 & 77.7 \\
\bottomrule
\end{tabular}
\end{adjustbox}
\par\vspace{2pt}{\scriptsize Top-1 accuracy (\%), ViT-B/16, CoOp splits; mean over seeds. Zero-shot CLIP does not depend on the shots. The linear probe, Tip-Adapter, Tip-Adapter-F, GDA and LP++ tune on a few-shot validation list of at most four images per class. CoOp is a published number (one run). Bold: best of each column.}
\endgroup
  \caption{\textbf{CLIP few-shot classification.}
  Top-1 accuracy (\%) with CLIP ViT-B/16 on the CoOp splits at 1, 4 and 16 shots per class.
  Mean over three seeds.
  The CoOp~\cite{zhou2022coop} row gives the published numbers.}
  \label{tab:supp-clip}
\end{table*}

\noindent\textbf{Results.}
The \ours{} reaches a higher accuracy than zero-shot CLIP on most sets, and a lower accuracy than the tuned adapters (Tab.~\ref{tab:supp-clip}): in no cell does it outperform all three of LP++, Tip-Adapter-F and GDA.
It scores closest to them at one shot, where it ties with LP++ on Pets (\opusOvlClipPetsNone{} against \lppFewshotClipPetsNone{}), and the gap widens with more shots: on EuroSAT at 16 shots, the \ours{} reaches \opusOvlClipEurosatNsixteen{} and GDA reaches \gdaFewshotClipEurosatNsixteen.
The programs that the agent writes for CLIP mostly add a scaled class-mean image feature to the zero-shot text row, with optional centering and a bias term.
We include the study because it tests the same harness on a third head and shows the limit of a program written from few images.

\section{CD-ViTO and CD-FSOD}
\label{sec:supp-J}

\subsection{CD-ViTO on our sets}
\label{sec:supp-J-setup}

CD-ViTO~\cite{fu2024cdvito} fine-tunes a detector built on DINOv2 ViT-L with a region-proposal network: it starts from a checkpoint trained on COCO, builds class prototypes from the support images, and fine-tunes the classification and box heads and a small adapter.
We run the official five-shot configuration of the authors' public repository and tune no hyperparameters on our sets; the tuned variant takes, for each set, the number of fine-tuning iterations and the learning rate that score best on the test images at seed 0 and reruns that setting on every seed, so it uses test labels in its favor.
Before we trusted any number on our sets, we reproduced the published result on Clipart1k: five-shot AP on the official test set is \cdvReproAp{}, against a published value of \cdfsodPubClipart{}.
All variants use the same training and test images as the other methods, with boxes only, so CD-ViTO is scored with box IoU on every set (the dagger in \Mtab{tab:main}), which differs from the scorer of the other rows only on pets.

\noindent\textbf{Excluded detectors.}
\label{sec:supp-J-notrun}
The detector of Yu et al.~\cite{yu2026ftfsod}, released as FT-FSOD, starts from a base model that was pretrained on Objects365, GoldG, V3Det, LVIS, GRIT and RefCOCO besides COCO, which puts it outside a COCO-source comparison.
PSP-FSOD~\cite{zhuo2026psp} has no public code.

\subsection{CD-FSOD benchmark}
\label{sec:supp-J-bench}

We use three target domains of CD-FSOD~\cite{fu2024cdvito}, ArTaxOr, Clipart1k and DIOR, with COCO as the source and five shots per class (Tab.~\ref{tab:supp-cdfsod}); Clipart1k uses the official test set and the official five-shot set, which the authors ship for one seed, and ArTaxOr and DIOR use our own splits.
The frozen masks of PMT set an upper bound on DIOR: only \dirCeiling{} of its objects get a proposal at IoU~0.5, and \dirOverCapShare{} of its instances sit in images that contain more objects than the 100 detections kept per image.
The Clipart1k classes are PASCAL VOC names that map one to one onto COCO rows, and the stock head already finds \stockCdfsodClipart{} of its objects with the right class.
On ArTaxOr and DIOR our edits lose \cdfsodOursDropLo--\cdfsodOursDropHi{} COCO points, against \cdfsodGradDropLo--\cdfsodGradDropHi{} for the unguarded heads (Prototype, LP and the cosine head).
Our edits score a lower box AP than the best published results, the highest five-shot AP per domain of CD-ViTO, FT-FSOD and PSP-FSOD (the row ``Best published of three'' in Tab.~\ref{tab:supp-cdfsod}).

\begin{table}[t]
  \centering
  \small
\begingroup\setlength{\tabcolsep}{3pt}
\begin{adjustbox}{max width=\linewidth}
\begin{tabular}{lcccccc}
\toprule
 & \multicolumn{3}{c}{Box AP $\times$100} & \multicolumn{3}{c}{COCO points lost} \\
\cmidrule(lr){2-4}\cmidrule(lr){5-7}
Method & ArTaxOr & Clipart1k & DIOR & ArTaxOr & Clipart1k & DIOR \\
\midrule
Stock head & 4.6 & 53.6 & 0.0 & 0.1 & 0.0$^\P$ & 0.2 \\
Prototype & 30.1\pmstd{1.7} & 36.7\pmstd{0.0} & \textbf{16.0}\pmstd{1.3} & 7.6\pmstd{9.3} & 6.5\pmstd{0.0}$^\P$ & 15.3\pmstd{15.7} \\
Linear probe (LP) & 21.2\pmstd{2.8} & 22.1\pmstd{0.7} & 12.8\pmstd{0.5} & 22.5\pmstd{18.2} & 33.2\pmstd{3.8}$^\P$ & 20.8\pmstd{6.6} \\
Cosine head & 17.2\pmstd{3.7} & 20.7\pmstd{1.9} & 13.9\pmstd{1.8} & 38.0\pmstd{20.6} & 31.1\pmstd{14.1}$^\P$ & 34.6\pmstd{22.7} \\
\oursF{} & \textbf{40.0}\pmstd{4.4} & 49.2\pmstd{0.0} & 11.3\pmstd{2.5} & \textbf{0.4}\pmstd{0.3} & 2.6\pmstd{0.0}$^\P$ & 0.9\pmstd{0.3} \\
\ours{} (Opus) & 32.4\pmstd{10.0} & \textbf{53.9}\pmstd{0.5} & 14.2\pmstd{3.0} & 1.2\pmstd{0.4} & \textbf{0.0}\pmstd{0.0}$^\P$ & \textbf{0.8}\pmstd{0.3} \\
\textcolor{gray}{Head fit on test images} & \textcolor{gray}{43.1} & \textcolor{gray}{47.4} & \textcolor{gray}{28.2} & \textcolor{gray}{14.7} & \textcolor{gray}{4.7$^\P$} & \textcolor{gray}{2.9} \\
\midrule
\textcolor{gray}{CD-ViTO, published} & \textcolor{gray}{47.9} & \textcolor{gray}{41.1} & \textcolor{gray}{26.9} & \na & \na & \na \\
\textcolor{gray}{Best published of three} & \textcolor{gray}{76.8} & \textcolor{gray}{60.5} & \textcolor{gray}{35.3} & \na & \na & \na \\
\bottomrule
\end{tabular}
\end{adjustbox}
\par\vspace{2pt}{\scriptsize CD-FSOD, five shots, box AP over IoU 0.5 to 0.95 (the boxes of the predicted masks), mean$\pm$std over seeds 0--2 (stock head and the head fit on test images: one seed). ArTaxOr and DIOR use our own splits, so the published rows are context, not a comparison: they are other detectors fine-tuned end to end. COCO lost: mask-mAP points on held-out COCO images, new rows joined to the 80 stock rows. $^\P$ Clipart1k: its 20 classes are COCO classes, so the edited rows replace their own COCO rows (in-place); the joined protocol would count every object twice. Bold: best in-head row per column; in the COCO columns, the lowest loss among the non-stock rows. Gray: not a few-shot method or another detector.}
\endgroup
  \caption{\textbf{CD-FSOD, five shots.}
  Box AP on three target domains and COCO mask mAP lost.
  The head fit on test images is a linear probe fit on the test images themselves, an upper reference and not a few-shot result.
  It and the stock head do not depend on the training draw, so each has one value; a dash marks a COCO score that the published detectors do not report.}
  \label{tab:supp-cdfsod}
\end{table}

\end{document}